\pdfoutput=1
\documentclass[11pt, letterpaper, logo, onecolumn, copyright, numbering]{paper_improved}

\usepackage{soul}
\usepackage{xcolor}

\usepackage{booktabs}
\usepackage{enumitem}
\usepackage{array}
\usepackage{colortbl}
\usepackage{xcolor}
\usepackage{caption}
\usepackage{tabularray}
\usepackage[utf8]{inputenc} 
\usepackage[T1]{fontenc}    
\usepackage{hyperref}       
\usepackage{url}            
\usepackage{booktabs}       
\usepackage{amsfonts}       
\usepackage{nicefrac}       
\usepackage{microtype}      
\usepackage{xcolor}         
\usepackage{graphicx}

\usepackage{natbib}
\usepackage{verbatim}
\usepackage{inconsolata}
\usepackage{tabularx}
\usepackage{float}
\usepackage{multirow}
\usepackage{listings}
\usepackage{xcolor}
\usepackage{subfigure}
\usepackage{subcaption}
\usepackage{amsmath}
\usepackage{amssymb}
\usepackage{array}
\usepackage{svg}
\usepackage[T1]{fontenc}
\definecolor{lightgray}{rgb}{0.95, 0.95, 0.95}
\definecolor{darkgray}{rgb}{0.4, 0.4, 0.4}
\definecolor{backcolour}{rgb}{0.95,0.95,0.92}
\definecolor{myblue}{rgb}{0.2, 0.4, 0.8} 
\definecolor{mygreen}{rgb}{0.2, 0.6, 0.2} 

\usepackage{amsmath,amssymb,amsfonts}

\usepackage{CJKutf8}
\usepackage{soul}
\usepackage{url}
\usepackage[utf8]{inputenc}
\usepackage{caption}
\usepackage{graphicx}
\usepackage{xcolor}
\usepackage{amsmath}

\usepackage{amsthm}
\usepackage{booktabs}
\usepackage{algorithm}
\usepackage{latexsym}
\usepackage{graphicx} 
\usepackage{amsmath}
\usepackage{subfigure}
\usepackage{microtype}
\usepackage{algorithmic}
\usepackage[switch]{lineno}
\usepackage[most]{tcolorbox}
\usepackage{alltt}

\definecolor{forestgreen}{rgb}{0.13, 0.55, 0.13}

\tcbset{
    aibox/.style={
        colback=white, 
        colframe=blue!75!black, 
        colbacktitle=blue!85!black, 
        coltitle=white, 
        fonttitle=\bfseries,
        enhanced,  
        drop shadow=black!50!white,  
        boxrule=1pt,  
        boxsep=10pt,  
        left=10pt,  
        right=10pt,  
        top=6pt,  
        bottom=6pt,  
        title code={\path[tcb fill frame] ([xshift=-10pt]frame.west) -- (frame.north west) -- (frame.north east) -- ([xshift=10pt]frame.east) -- cycle;},  
        attach boxed title to top left={xshift=10pt, yshift*=-\tcboxedtitleheight/2},
        boxed title style={boxrule=0pt, frame code={}}  
    }
}

\newtcolorbox{AIbox}[2][]{aibox, title=#2, #1}

\let\cite\citep

\title{New Trends for Modern Machine Translation with Large Reasoning Models}

\reportnumber{} 

\author[*,1]{Sinuo Liu\textsuperscript{1,2}, Chenyang Lyu\textsuperscript{1}, Minghao Wu\textsuperscript{1}, Longyue Wang\textsuperscript{1}, Weihua Luo\textsuperscript{1}, Kaifu Zhang\textsuperscript{1}, Zifu Shang\textsuperscript{1}\\ ~ \\ 
\bf 1. MarcoPolo Team, Alibaba International Digital Commerce\\
\bf 2. University of Edinburgh}

\begin{abstract}
Recent advances in Large Reasoning Models~(LRMs), particularly those leveraging Chain-of-Thought reasoning~(CoT), have opened brand new possibility for Machine Translation~(MT). This position paper argues that LRMs substantially transformed traditional neural MT as well as LLMs-based MT paradigms by reframing translation as a dynamic reasoning task that requires contextual, cultural, and linguistic understanding and reasoning. We identify three foundational shifts: 1) \textit{contextual coherence}, where LRMs resolve ambiguities and preserve discourse structure through explicit reasoning over cross-sentence and complex context or even lack of context; 2) \textit{cultural intentionality}, enabling models to adapt outputs by inferring speaker intent, audience expectations, and socio-linguistic norms; 3) \textit{self-reflection}, LRMs can perform self-reflection during the inference time to correct the potential errors in translation especially extremely noisy cases, showing better robustness compared to simply mapping \textit{X->Y} translation. We explore various scenarios in translation including stylized translation, document-level translation and multimodal translation by showcasing empirical examples that demonstrate the superiority of LRMs in translation. We also identify several interesting phenomenons for LRMs for MT including auto-pivot translation as well as the critical challenges such as over-localisation in translation and inference efficiency. In conclusion, we think that LRMs redefine translation systems not merely as text converters but as multilingual cognitive agents capable of reasoning about meaning \textit{beyond the text}. This paradigm shift reminds us to think of problems in translation beyond traditional translation scenarios in a much broader context with LRMs - what we can achieve on top of it. 

\end{abstract}
\begin{document}
\begin{CJK}{UTF8}{gbsn}

\maketitle

\begin{figure*}[h]
    \centering
    \includegraphics[width=0.97\linewidth]{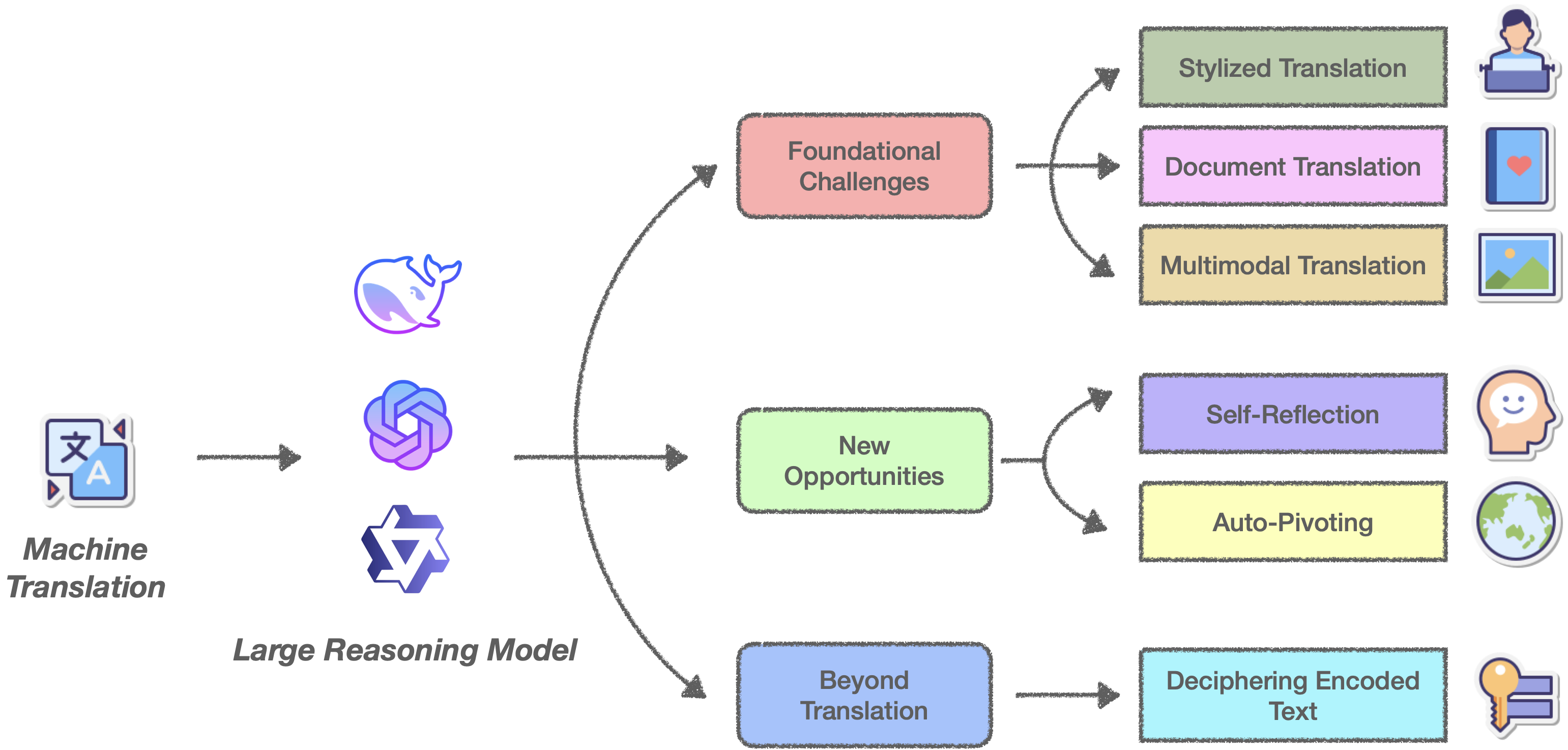}
    \caption{Promsing directions for MT using LRMs (e.g., DeepSeek R1), including some foundational and classical MT scenarios such as stylized translation, new challenges with LRMs like self-reflection, and some new challenges for LRMs.}
    \label{fig:directions}
\end{figure*}

\section{Introduction}

As a fundamental component of Natural Language Processing~(NLP), Machine Translation~(MT) enables cross-linguistic communication by automatically converting text between different languages. ~\cite{tsujii1986future_MT,DBLP:conf/coling/SatoN90_MT}. As globalization accelerates, the demand for accurate and efficient translation systems has grown exponentially, making MT a cornerstone of modern NLP research and applications. The introduction of Neural Machine Translation~(NMT) marked a significant leap forward in the field. By leveraging deep learning techniques, NMT systems have demonstrated the ability to capture complex linguistic patterns and contextual dependencies, significantly improving translation quality compared to earlier approaches~\cite{Vaswani17attention,castilho2017neural,stahlberg2020neural,kocmi-etal-2022-findings}. However, despite these advancements, NMT systems still face challenges such as translating idiomatic expressions, handling low-resource languages, and maintaining coherence across long documents~\cite{koehn2017six,wang2019discourse,yang2020survey-mt,haddow-etal-2022-survey-mt}. These limitations highlight the need for more robust and adaptive translation systems.

The emergence of Large Language Models~(LLMs), such as GPT-3, GPT-4, LLaMA, Qwen and many others~\cite{brown2020language-gpt3,chen2021evaluating-codex-gpt3,ouyangtraining-instructGPT,weifinetuned,hadi2023large,touvron2023llama1,qwen2025qwen25technicalreport}, has further revolutionized MT. Unlike traditional NMT systems that rely on extensive parallel corpora, LLMs excel in zero-shot and few-shot translation scenarios, often achieving performance comparable to supervised systems~\cite{jiao2023chatgpt,robinson2023chatgpt_mt,moslem2023adaptive,pang2024salute,lyu-etal-2024-paradigm-mt-llms,zhang2025exploringtranslationmechanismlarge}. Beyond their translation capabilities, LLMs have demonstrated remarkable versatility in tasks such as style transfer, summarization, and question answering~\cite{bang2023multitask_eval_chatgpt,laskar-etal-2023-systematic_eval_chatgpt,li2023bactrian}, opening new avenues for MT research~\cite{he2023exploring_maps_mt,he2024improving_mt}. However, LLMs also introduce challenges, such as privacy concerns and the need for interpretability in their decision-making processes~\cite{klymenko-etal-2022-differential,privatenlp-2022-privacy-nlp,li2023privacypreserving}.

Building on the success of LLMs, the development of Large Reasoning Models~(LRMs)~\cite{jaech2024openai_o1,zhao2024marco_o1,qwq-32b-preview,deepseekai2025deepseekr1incentivizingreasoningcapability}represents the next evolution in MT. LRMs integrate reasoning capabilities, such as Chain-of-Thought~(CoT) reasoning~\cite{wei2022chain_cot}, enabling them to tackle translation as a dynamic reasoning task. This approach allows LRMs to address challenges like contextual coherence, cultural intentionality, and compositional generalization, making them more robust and interpretable than traditional LLMs~\cite{wang2024drt_o1,chen2025evaluating_o1_mt}. For example, LRMs exhibit self-reflection capabilities, allowing them to correct errors during inference, particularly in noisy or ambiguous cases.


In this position paper, we explore the transformative potential of LRMs in redefining MT systems. By leveraging CoT reasoning, LRMs reframe translation as a dynamic reasoning task that goes beyond traditional text-to-text mapping, requiring deep contextual, cultural, and linguistic understanding. We identify three foundational shifts brought by LRMs: 1) \textit{contextual coherence}, where LRMs resolve ambiguities and preserve discourse structure through explicit reasoning over cross-sentence and complex contexts, even in cases of limited or noisy input; 2) \textit{cultural intentionality}, enabling models to adapt translations by inferring speaker intent, audience expectations, and socio-linguistic norms; and 3) \textit{self-reflection}, where LRMs can iteratively refine translations during inference, correcting errors and demonstrating superior robustness in challenging scenarios. These capabilities position LRMs as a significant advancement over both traditional neural MT and LLMs-based approaches.

We investigate various translation scenarios to demonstrate the superiority of LRMs, including stylized translation~\cite{wang2022controlling,Sennrich16controlling}, document-level translation~\cite{wang2024findings_wmt_24}, and multi-modal translation~\cite{sulubacak2020multimodal}. Through empirical case examples, we showcase how LRMs show impressive capability in tasks such as preserving stylistic features, maintaining consistency across long documents, and integrating visual context for multi-modal inputs. Additionally, we identify various interesting phenomena in LRMs for translation, such as \textit{auto-pivot translation}, where LRMs automatically used English/Chinese as the pivotal language to bridge the translation between two other languages without explicit instructions, and \textit{over-localization}, a challenge where models may over-adapt translations to local norms at the expense of global coherence. We also discuss critical challenges such as inference efficiency~\cite{xia2025tokenskipcontrollablechainofthoughtcompression}, which remains a key problem for optimization as LRMs scale to more complex tasks.

This position paper argues that LRMs redefine MT systems not merely as text converters but as multilingual cognitive agents capable of reasoning about meaning beyond the text. By enabling translation systems to dynamically reason about context, culture, and intent, LRMs open up new possibilities for translation with its superior reasoning capability. We conclude by highlighting the opportunities and challenges for future research, including the need to address over-localization, improve inference efficiency, and explore the broader implications of LRMs in rethinking translation as a reasoning-driven task. This paradigm shift invites us to envision translation not just as a linguistic challenge but as a gateway to deeper cross-cultural understanding and communication.

\section{Foundational Challenges in MT for LRMs}

In this section, we explore how LRMs when dealing with challenges that have plagued MT from the past to nowadays. We focus on two classical yet critical tasks: stylized translation, document-level translation, and the use of multi-modal reasoning with translation. These challenges have historically served as benchmarks for evaluating the capabilities of MT systems, and we demonstrate how LRMs, equipped with reasoning abilities, offer innovative solutions while also revealing new complexities.

\subsection{Stylized Translation}

\begin{figure}[!ht]
\centering
\begin{CJK}{UTF8}{gbsn}
\scalebox{0.8}{
\begin{AIbox}{Stylized Translation}
\small{\bf Prompt:} Please translate the following text into Chinese. Analyze the style of the originla text and translate following the style you have analyzed:
荒海や　佐渡によこたふ　天河\\

{\bf Output:}\\

\textbf{\textit{DeepSeek-R1}}

\textcolor{blue}{怒海接佐渡，天河横空流}  \textit{(generation 1)}\\
\textcolor{blue}{怒涛涌佐渡，天河横亘贯长空}  \textit{(generation 2)}\\
\textcolor{blue}{怒涛涌荒海，佐渡孤横亘，天河悬夜空}  \textit{(generation 3)}

\textbf{\textit{DeepSeek-V3}}

\textcolor{blue}{荒海阔　横跨佐渡　银河落}

\textbf{\textit{OpenAI-o3-mini}}

\textcolor{blue}{荒海上，佐渡横卧处，天河遥望。}

\textbf{\textit{QwQ-32B-Preview}}

\textcolor{blue}{狂海呀，佐渡横，银河。}

\tcbline

\small{\bf Prompt:} Please translate the following Japanese Haiku into Chinese following Haiku format:
\textcolor{blue}{荒海や 佐渡によこたふ 天河} \\

{\bf Output:}\\

\textbf{\textit{DeepSeek-R1}}

\textcolor{blue}{怒海接苍茫，\\
佐渡横悬天汉淌，\\
星涛夜未央。}

\hfill
\end{AIbox}
}
\end{CJK}
\caption{An example of prompting DeepSeek-R1 to translate a Japanese Haiku into Chinese and following Haiku format. In the above case, when we tell the model to analyze the format of the original text first and generate the translation with the format which is analyzed by the model itself, R1 model will usually generate an over-localized translation and does not follow the 5-7-5 format of Haiku. Other models also does nto generate the 5-7-5 format, but they will at least generate a three line translation. However, in the lower case, if we tell the R1 model to generate translation following Haiku format, R1 model will usually generate the translation following the Haiku 5-7-5 format.}	
\label{fig:stylized_mt_example}
\end{figure}

Stylized translation involves generating translations that preserve the stylistic features of the source text, such as tone, formality, or genre-specific expressions. Traditional MT systems often rely on multi-parallel datasets or post-processing techniques like style transfer to achieve this~\cite{Niu20controlling,wang2022controlling}. While LLMs have simplified stylized translation through natural language prompts, their performance can be inconsistent in zero-shot scenarios. Without explicit instructions, LLMs may fail to analyze the stylistic nuances of the source text, leading to translations that lack cultural or contextual appropriateness~\cite{lyu-etal-2024-paradigm-mt-llms}.


With LRMs, the model will first figure out the scenario of the original text. With such reasoning, the model will have a better understanding of the style of translation should it apply to the translated output. The reasoning ability also provides the models with a better performance in understanding the idea of the original test, understanding what the original text wants to express, and choosing suitable words for the generated output without changing what the original text wants to express. However, this approach can sometimes lead to \textit{over-localization}, where the translation adheres more closely to the target language's norms than to the source text's original style. By asking the LRMs to analyze the format and style of the original text and generate an output which follows the format and style it has analyzed with the original text, the LRMs are more likely to generate an output which does not follow the style of the original text, but rather generate an translation more closer to the style of the target language. Figure~\ref{fig:stylized_mt_example} illustrates this phenomenon with a Japanese Haiku translated into Chinese. While DeepSeek R1 successfully adapts the Haiku to a Chinese poetic format, it occasionally deviates from the strict 5-7-5 syllable structure, favoring patterns more familiar to Chinese readers. This raises important questions about the trade-off between preserving the source text's authenticity and ensuring the translation's accessibility and appeal in the target language. Additional examples, such as translations of Spanish poetry, are provided in the appendix.


\begin{figure}[!ht]
\centering
\begin{CJK}{UTF8}{gbsn}
\scalebox{0.8}{
\begin{AIbox}{Stylized Translation}
\small{\bf Prompt:} Translate the text from Chinese to English: 预售官方正版哪吒周边电影《哪吒之魔童闹海》艺术家系列徽章盲袋\\

{\bf Output:}\\

\textbf{\textit{DeepSeek-R1-14B}}
<think>
Alright, let's break down this translation request. The user has provided a Chinese sentence that needs to be translated into English. The original text is about 
pre-sale official merchandise related to the movie "Nezha's Demon Child Turmoil at Sea." It mentions an artist series badge blind bag.

First, I'll look at each part of the Chinese text:
- 预售: This means "pre-sale" in English.
- 官方正版: This translates to "official genuine version" or simply "official."
- 哪吒周边: Refers to "Nezha周边," which is merchandise related to Nezha. "周边" here can be translated as "merchandise" or "goods."
- 电影: This means "movie."
- 《哪吒之魔童闹海》: The title of the movie, which I've previously translated as "Nezha's Demon Child Turmoil at Sea." For consistency, I'll use the same 
translation.
- 艺术家系列徽章盲袋: This is a bit more complex. 
  - 艺术家系列: Means "artist series."
  - 徽章: Translates to "badge" or "pin."
  - 盲袋: Refers to "blind bag," which is a type of packaging where the contents are randomly assorted.

Putting it all together, the sentence can be translated as:
"Pre-sale official Nezha movie merchandise 'Nezha's Demon Child Turmoil at Sea' artist series badge blind bags."

I need to ensure that the translation captures both the promotional aspect (pre-sale) and the nature of the product (artist series badges in a blind bag format). 
It's important to maintain the official status and tie it to the movie title for clarity.

Additionally, considering the context might be marketing or retail related, the translation should appeal to potential buyers interested in collectibles. Therefore, 
using terms like "merchandise" and "blind bags" makes sense as they are commonly used in that industry.

I also need to make sure that the structure is clear and concise, avoiding overly complicated phrasing while ensuring all key elements are included.
</think>

\textbf{\textcolor{blue}{Pre-sale official Nezha movie merchandise 'Nezha's Demon Child Turmoil at Sea' artist series badge blind bags}}

\hfill
\end{AIbox}
}
\end{CJK}
\caption{Translation of a Chinese product description into English using different models, showcasing the reasoning process of DeepSeek R1.}	
\label{fig:stylized_mt_example_2}
\end{figure}

\subsection{Document-Level Translation}


Document-level translation~\cite{wang2023document,wang2024findings_wmt_24} has long been a challenge for MT systems due to the complexity of maintaining consistency, coherence, and stylistic integrity across lengthy texts. Issues such as keyword unification, pronoun resolution, and tone consistency are particularly problematic. While LLMs have made significant strides in this area, LRMs further enhance document-level translation by leveraging their reasoning capabilities to better understand and unify context across sentences and paragraphs.

\begin{figure}[!t]
\centering
\begin{CJK}{UTF8}{gbsn}
\resizebox{0.8\textwidth}{!}{
\begin{AIbox}{Document-Level Translation}
{\bf Prompt:} \\
\small Please translate the following text into Chinese.\\
This study aims to make use of two concepts in the field of aeroacoustics; an analogy with relativity, and Geometric Algebra. The analogy with relativity has been investigated in physics and cosmology, but less has been done to use this work in the field of aeroacoustics. Despite being successfully applied to a variety of fields, Geometric Algebra has yet to be applied to acoustics. Our aim is to apply these concepts first to a simple problem in aeroacoustics, sound propagation in uniform flow, and the more general problem of acoustic propagation in non-uniform flows. By using Geometric Algebra we are able to provide a simple geometric interpretation to a transformation commonly used to solve for sound fields in uniform flow. We are then able to extend this concept to an acoustic space- time applicable to irrotational, barotropic background flows. This geometrical framework is used to naturally derive the requirements that must be satisfied by the background flow in order for us to be able to solve for sound propagation in the non-uniform flow using the simple wave equation. We show that this is not possible in the most general situation, and provide an explicit expression that must be satisfied for the transformation to exist. We show that this requirement is automatically satisfied if the background flow is incompressible or uniform, and for both these cases derive an explicit transformation. In addition to a new physical interpretation for the transformation, we show that unlike previous investigations, our work is applicable to any frequency.\\

{\bf Output:} \\

\textit{\textbf{DeepSeek-R1}}

<think>
{\em{\small 好的，我现在需要将一段英文的学术摘要翻译成中文。首先，我要通读一遍原文，理解整体内容和主要概念。这段文字涉及气动声学中的两个概念：相对论类比和几何代数。作者提到他们首次将这些概念应用于气动声学中的简单问题，比如均匀流中的声传播，以及更复杂的非均匀流中的声传播问题。

接下来，我要注意一些专业术语的准确翻译。例如，“aeroacoustics”应译为“气动声学”，“relativity”是“相对论”，“Geometric Algebra”是“几何代数”。“Barotropic background flows”可能需要译为“正压背景流动”，“irrotational”是“无旋的”。还有一些技术性的表达，比如“wave equation”是“波动方程”，“incompressible”是“不可压缩的”。

然后，要确保句子的结构在中文中通顺自然。例如，原文中的长句可能需要拆分成更短的句子，或者调整语序以符合中文的表达习惯。比如，“By using Geometric Algebra we are able to provide...”可以翻译为“通过运用几何代数，我们能够为…提供一个简单的几何解释”。

还要注意保持原文的逻辑关系和学术严谨性。比如，“We show that this is not possible in the most general situation...”要准确传达出研究结果，即“在一般情况下这是不可能的，并给出了变换存在的明确条件”。

此外，有些句子中的连接词需要处理得当，比如“despite being successfully applied to...”译为“尽管已成功应用于多个领域，但几何代数尚未应用于声学领域”，这样既保留了对比关系，又符合中文表达。

最后，要检查整个译文的流畅性和一致性，确保专业术语前后统一，句子之间衔接自然。例如，“acoustic space-time”可以译为“声学时空”，保持术语的准确性。同时，注意段落的分隔，使译文结构清晰，层次分明。

完成初稿后，再通读一遍，对照原文确认没有遗漏或误译的地方。特别是技术细节部分，如“背景流动必须满足的条件”和“变换存在的显式表达式”等，确保信息准确无误。这样，最终的译文才能既忠实于原意，又符合中文学术写作的规范。}}
</think>\\

\textbf{\small Final output:}

\small 本研究旨在运用气动声学领域的两个概念：相对论类比与几何代数。相对论类比在物理学与宇宙学领域已得到深入研究，但在气动声学中的应用尚显不足。尽管几何代数已在多个领域成功应用，但其在声学领域的应用仍属空白。我们的目标是将这些概念首先应用于气动声学中的基础问题——均匀流中的声传播，继而拓展至更具普遍性的非均匀流声传播问题。

\small 通过运用几何代数，我们能够为均匀流声场求解中常用的变换方法提供简洁的几何解释。随后，我们将这一概念拓展至适用于无旋、正压背景流动的声学时空框架。借助该几何框架，我们自然推导出背景流动必须满足的条件，以便能通过简单波动方程求解非均匀流中的声传播问题。研究表明，在普遍情况下这一目标无法实现，并给出了变换存在的显式条件。我们证明该条件在背景流动不可压缩或均匀时自动成立，并针对这两种情形推导了显式变换公式。除赋予变换新的物理解释外，本研究还突破前人研究的局限，证明了所提方法适用于任意频率情形。
\hfill
\end{AIbox}
}
\end{CJK}
\caption{An example of translating long document text. Reasoning model will first extract all the keywords from the document to unify the translation of these words before starting to translate the text. Compare to the translation result given by DeepSeek V3, result given by R1 is more formal and close to a research paper writing style. The original text came from \cite{li2020docbank}.}
\label{fig:example_document_level_translation_R1}
\end{figure}

\begin{figure}
    \centering
    \begin{CJK}{UTF8}{gbsn}
    \scalebox{0.8}{
    \begin{AIbox}{Document-Level Translation}

    {\bf Output:} \\

    \textit{\textbf{DeepSeek-V3}}

    \small 本研究旨在利用航空声学领域中的两个概念：与相对论的类比以及几何代数。与相对论的类比在物理学和宇宙学中已有所探讨，但在航空声学领域中的应用却较少。尽管几何代数已成功应用于多个领域，但尚未涉足声学领域。我们的目标是将这些概念首先应用于航空声学中的一个简单问题——均匀流中的声传播，以及更为普遍的非均匀流中的声传播问题。通过运用几何代数，我们能够为常用于求解均匀流中声场的变换提供一个简洁的几何解释。随后，我们将这一概念扩展至适用于无旋、正压背景流的声学时空。这一几何框架自然地导出了背景流必须满足的条件，以便我们能够使用简单的波动方程求解非均匀流中的声传播。我们证明，在最一般情况下这是不可能的，并给出了变换存在所需满足的明确表达式。我们指出，若背景流不可压缩或均匀，则此条件自动满足，并为这两种情况推导出了明确的变换。除了为变换提供新的物理解释外，我们还表明，与以往研究不同，我们的工作适用于任意频率。
    
    \hfill
    \end{AIbox}
    }
    \end{CJK}
    \caption{Translation result given by DeepSeek V3, the prompt is the same as Figure \ref{fig:example_document_level_translation_R1}.}
    \label{fig:example_document_level_translation_V3}
\end{figure}


For instance, LRMs show strong ability at identifying and consistently translating key terms, resolving ambiguous pronouns, and adapting the tone of the translation to suit the target language's conventions. Figure~\ref{fig:example_document_level_translation_R1} demonstrates this capability with the translation of a scientific abstract. DeepSeek R1 not only maintains consistency in terminology but also adapts the tone to match the formal style expected in Chinese academic writing. In contrast, Figure~\ref{fig:example_document_level_translation_V3} highlights the limitations of non-reasoning models, which often overuse certain phrases (e.g., "我们") and fail to adhere to the target language's stylistic norms. The ability of LRMs to reason about context at the document level also enables them to handle complex narrative structures, such as those found in novels or legal documents. By understanding the relationships between sentences and paragraphs, LRMs can generate translations that preserve the logical flow and coherence of the original text. This represents a significant advancement over traditional MT systems, which often struggle with long-range dependencies and contextual ambiguities.

\subsection{Multi-Modal Reasoning with Translation}



\begin{figure}[]
\centering
\begin{CJK}{UTF8}{gbsn}
\scalebox{0.8}{
\begin{AIbox}{Multi-Modal Translation - Contextual Reasoning}
\small{\bf Prompt:} \\
\includegraphics[width=0.5\linewidth]{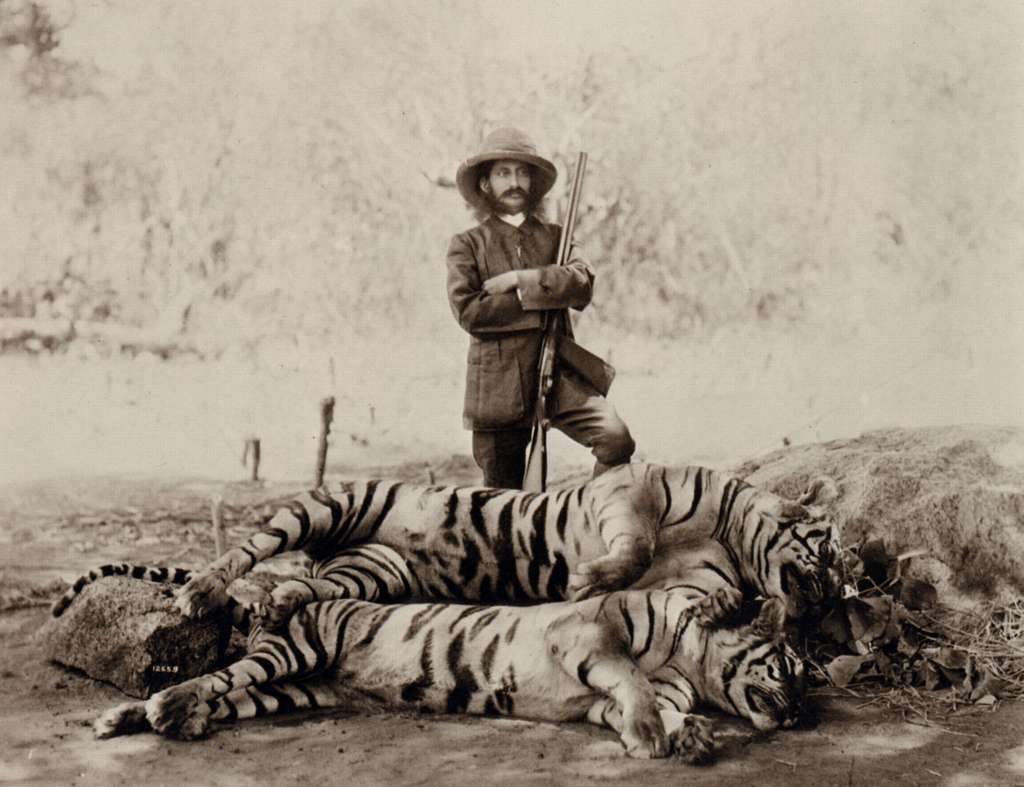}\\
请基于图片，将以下文本翻译成英文: "捕猎的是老虎"

{\bf Output:} \\

\textit{\textbf{QvQ-72B-preview}}
<think>
{\em{\small So I've got this black and white photograph here, and it's pretty old-looking, which makes sense given the context. There's a man standing behind two tigers that are lying on the ground. He's holding a rifle, and he's dressed in what appears to be hunting gear from a bygone era. The background is a bit blurred, but it looks like a natural setting, maybe a rocky or sandy area with some vegetation.

First off, I need to make sure I understand what's going on here. It seems like this man is posing with the tigers he has hunted, which unfortunately was a practice in some parts of the world historically. Tigers are large, powerful cats known for their distinctive orange fur with black stripes, and they're native to various regions in Asia.

Given that the photo is in black and white and the style of clothing the man is wearing, it's likely that this photograph was taken in the late 19th or early 20th century. During that time, big game hunting was a popular pastime among certain social classes, especially in colonial contexts.

The man is standing confidently, with one hand on his hip and the other holding the rifle. His attire includes a wide-brimmed hat, which was common for protection from the sun during outdoor activities. He's also wearing a jacket and trousers, which were typical for hunting expeditions to keep safe from the elements and wildlife.

Looking at the tigers, they are both lying on their sides, appearing lifeless, which unfortunately confirms that they have been hunted. Their stripes are clearly visible, even in black and white, showing the classic tiger pattern. The way they are positioned suggests they were arranged for the photograph to showcase the hunt's成果.

In the background, there are some indistinct features that might be rocks or small trees, but it's hard to make out any specific details due to the focus being on the man and the tigers in the foreground.

Overall, this image evokes a sense of history and raises questions about the ethics of hunting, particularly of endangered species like tigers. It's a stark reminder of how human activities have impacted wildlife populations over time.

Now, to answer the question: "捕猎的是老虎" translates to "The animal being hunted is a tiger." Given the context of the photograph, this statement is accurate. The man is clearly posing with two tigers, which he appears to have hunted, as evidenced by their lifeless state and his possession of a rifle.

However, it's important to note that in modern times, hunting tigers is strictly prohibited due to their endangered status. Conservation efforts are now focused on protecting these magnificent animals and their habitats to ensure their survival.

But based on the photograph provided, which seems to be from a historical period when such practices were unfortunately common, the translation "The animal being hunted is a tiger" is correct in describing the scene depicted.

So, to sum it up, the answer is:

Yes, the translation "The animal being hunted is a tiger" is accurate based on the photograph provided.}}
</think>

\textbf{\textit{Final answer: Yes, the translation \textcolor{blue}{“The animal being hunted is a tiger”} is accurate based on the photograph provided.}}

\hfill
\end{AIbox}
}
\end{CJK}
\caption{An example of asking the reasoning model to translate an ambiguous sentence and providing an image as the context. It shows that the model could translate correctly with the provided image.}	
\label{fig:example_contextual_commonsense_translation_1}
\end{figure}

\begin{figure}[]
\centering
\begin{CJK}{UTF8}{gbsn}
\scalebox{0.8}{
\begin{AIbox}{Multi-Modal Translation - Contextual Reasoning}
\small{\bf Prompt:} \\
\includegraphics[width=0.5\linewidth]{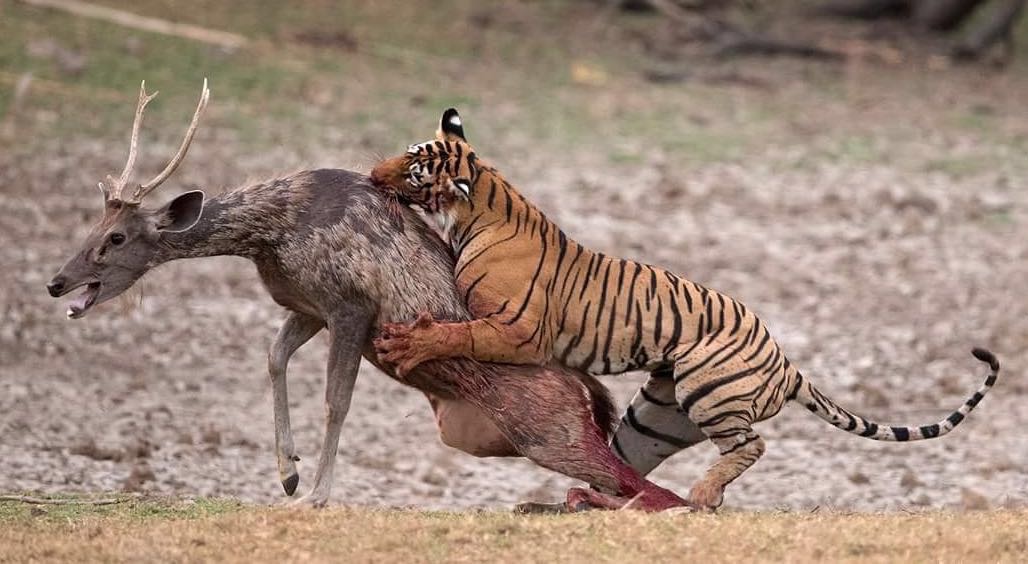}\\
请基于图片，将以下文本翻译成英文: "捕猎的是老虎"

{\bf Output:} \\

\textit{\textbf{QvQ-72B-preview}}

<think>
{\em{\small (Unable to provide as it's too long)}}
</think>

\textbf{\textit{Final Answer: \textcolor{blue}{It is the tiger that is hunting.}}}

\hfill
\end{AIbox}
}
\end{CJK}
\caption{Another example of translating an ambiguous sentence with image context. With the same prompt as Figure \ref{fig:example_contextual_commonsense_translation_1} but in different context, the model successfully understood the image context and translated the sentence correctly.}	
\label{fig:example_contextual_commonsense_translation_2}
\end{figure}

The integration of multi-modal inputs, such as images, videos, or audio, has expanded the capabilities of MT systems. LLMs like GPT-4o~\cite{hurst2024gpt4o} have demonstrated the ability to process and translate text in conjunction with visual or auditory context, enabling more accurate and context-aware translations. For example, when translating a sentence like "他在看报纸," an accompanying image can help the model determine whether the subject is reading a physical newspaper or browsing a digital one. This multi-modal approach allows LLMs to resolve ambiguities that are difficult to address with text alone. LRMs further enhance multi-modal translation by incorporating reasoning capabilities into the process~\cite{qvq-72b-preview}. Unlike traditional LLMs, which primarily rely on pattern recognition, LRMs can infer relationships between textual and non-textual inputs, enabling deeper contextual understanding. For instance, when presented with an ambiguous sentence and an accompanying image, LRMs can reason about the visual context to generate translations that align with the intended meaning. Figures~\ref{fig:example_contextual_commonsense_translation_1} and~\ref{fig:example_contextual_commonsense_translation_2} illustrate how LRMs leverage visual context to disambiguate translations, demonstrating their superior ability to handle complex multi-modal scenarios.

\begin{figure}[h!]
    \centering
    \begin{CJK}{UTF8}{gbsn}
    \scalebox{0.8}{
    \begin{AIbox}{Multi-Modal Translation: Sign Language Translation}
        \small{\bf Prompt:}\\
        \includegraphics[width=0.25\linewidth]{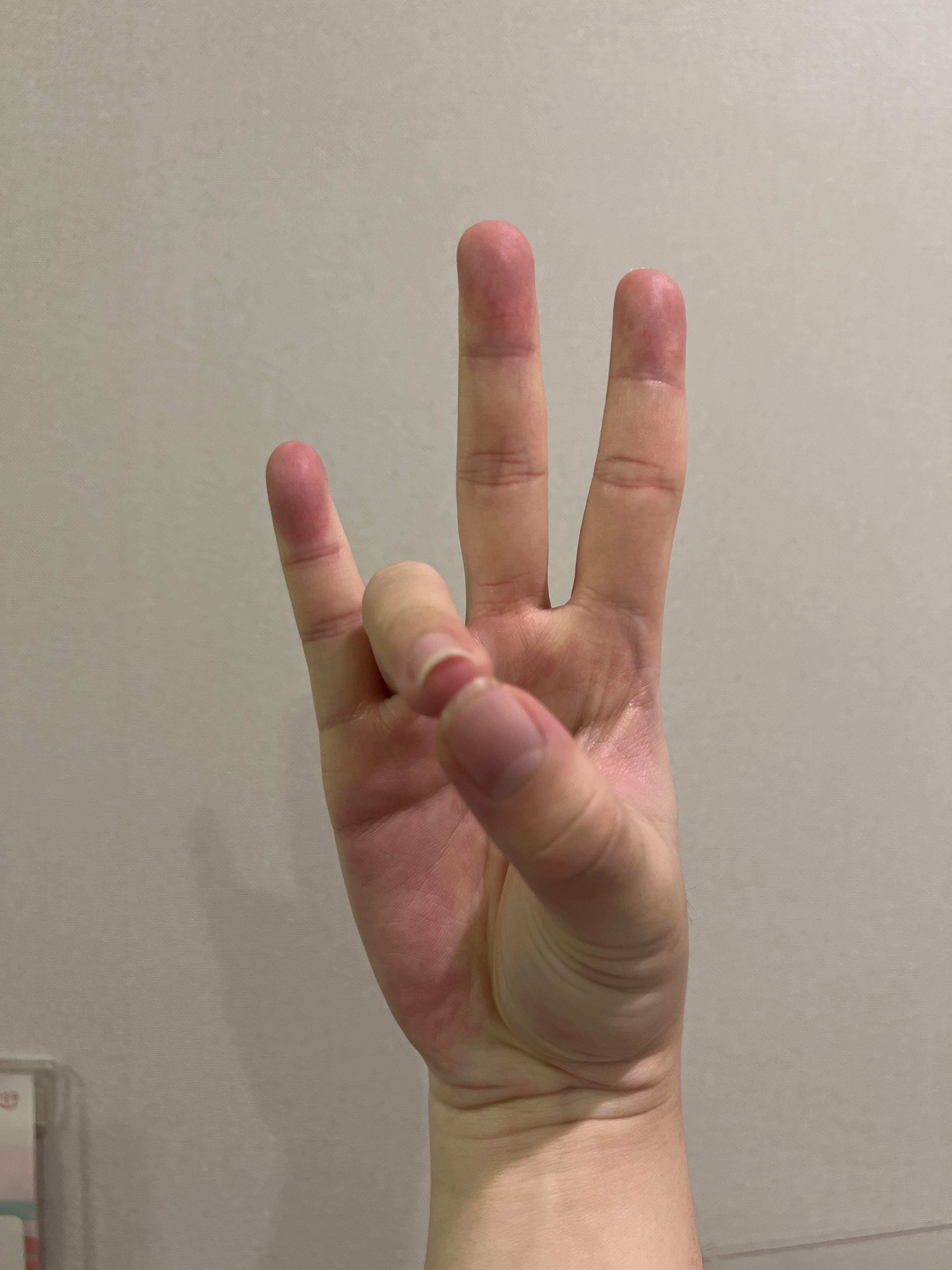}\\
        Which number does this sign shows in American Sign Language?

        {\bf Output:}

        \textbf{\textit{OpenAI-o3-mini}}

        The hand sign shown in the image corresponds to the number "\textbf{\textcolor{blue}{3}}" in American Sign Language (ASL). This is made by extending the index, middle, and thumb fingers while keeping the other fingers folded down.

        \textbf{\textit{QVQ-72B-Preview}}

        Final Answer: \textbf{\textcolor{blue}{4}}

        \hfill
    \end{AIbox}
    }
    \end{CJK}
    \caption{An example of prompting the OpenAI GPT-3o-mini to translate American Sign Language numbers into Arabic numbers. However, the model failed to understand the image.}
    \label{fig:ASL_7_translate_example}
\end{figure}

However, challenges remain in handling specialized multi-modal inputs, such as sign language or highly domain-specific visual content. Figure~\ref{fig:ASL_7_translate_example} shows an example where OpenAI-o3-mini fails to interpret an American Sign Language gesture from an image. This highlights the limitations of current multi-modal reasoning capabilities, particularly in areas requiring fine-grained understanding of non-textual inputs. To address these challenges, future research could explore the integration of domain-specific knowledge, such as sign language dictionaries or gesture recognition algorithms, into LRMs. Additionally, advancements in multi-modal training datasets and architectures could further enhance the ability of LRMs to process and translate complex multi-modal inputs effectively.



\section{New Challenges and Opportunities with Reasoning-Enhanced MT}


As LRMs introduce reasoning capabilities to MT, they also bring new challenges and opportunities. In this section, we explore several new characteristics of LRMs in MT such as self-reflection and the use of intermediate language during translation, while also identifying areas for further improvement.

\subsection{Self-Reflection}



One of the key advantages of LRMs is their ability to perform self-reflection during the translation process~\cite{deepseekai2025deepseekr1incentivizingreasoningcapability}. This allows them to identify and correct errors, particularly in ambiguous or noisy input scenarios ike when there are typos existed in the input, or the input sentence has been randomly rearranged into a sentence which could not be read normally. For example, when translating an ambiguous Chinese sentence like "捕获的是猎人," DeepSeek R1 initially interprets it as "The hunter is the one who hunts." However, through self-reflection, the model revisits its reasoning and considers an alternative interpretation: "The one who captures is the hunter." This iterative process demonstrates the potential of LRMs to refine translations dynamically, though further research is needed to fully understand the scope and limitations of this capability.

Self-reflection also enables LRMs to handle noisy or incomplete input more effectively. For instance, when presented with a sentence containing typos or grammatical errors, LRMs can infer the intended meaning and generate a coherent translation. This capability is particularly valuable in real-world applications, where input quality can vary significantly. However, the effectiveness of self-reflection depends on the model's ability to accurately assess its own reasoning process, which remains an area of active research.

\subsection{Auto-Pivot Translation}

An interesting behavior observed in LRMs is their automatic use of a pivot or bridge language during the translation process, even without explicit instruction~\cite{wu2007pivot,kim-etal-2019-pivot,dabre2021simultaneousmultipivotneuralmachine}. This phenomenon, which we call \textit{auto-pivot translation}, occurs when LRMs internally reason through a high-resource language (such as English or Chinese) to translate between less-resourced languages. For example, when translating from Irish to Chinese, an LRM might first translate the Irish text into English and then generate the final Chinese translation based on the English intermediate output. This behavior is evident in the reasoning chains of LRMs, as shown in Figure~\ref{fig:example_minor_language_translation_intermediate_language}.

The auto-pivot translation phenomenon highlights the model's reliance on high-resource languages as a scaffolding mechanism for reasoning. This approach leverages the extensive training data and linguistic knowledge available for high-resource languages, enabling LRMs to handle low-resource language pairs more effectively. However, it also raises important questions about the transparency and efficiency of the translation process. For instance, the automatic insertion of an intermediate language step may introduce additional computational overhead and potential distortions, particularly when the pivot language lacks equivalent expressions. One of the key implications of auto-pivot translation is its impact on translation quality and cultural fidelity. While the use of a pivot language can improve fluency and coherence, it may also lead to inaccuracies or loss of meaning, especially for idiomatic expressions or culturally specific terms. For example, a proverb in Irish might lose its cultural significance when first translated into English and then into Chinese. Additionally, the choice of pivot language can influence the final output, as different high-resource languages may capture different aspects of the source text's meaning.


\begin{figure}[]
\centering
\begin{CJK}{UTF8}{gbsn}
\scalebox{0.8}{
\begin{AIbox}{Auto-pivot Translation}
\small{\bf Prompt:} Please translate the following Irish sentence into Chinese: Ní leor teanga amháin\\

{\bf Output:} 

\textbf{\textit{DeepSeek-R1}}

<think>
{\em{\small Okay, I need to translate the Irish sentence "Ní leor teanga amháin" into Chinese. Let me start by breaking down the sentence word by word.

First, "Ní" is a negative particle in Irish, so it means "not". Then "leor" translates to "enough" or "sufficient". "Teanga" means "language", and "amháin" is "one" or "single". Putting that together, the literal translation would be "Not enough language one" or more naturally, \textcolor{red}{"One language is not enough."}

Now, I need to make sure that the translation into Chinese is both accurate and natural. The user probably wants a concise and correct translation. The phrase "One language is not enough" emphasizes the insufficiency of a single language. In Chinese, a common way to express this might be “掌握一种语言是不够的” which translates back to "Mastering one language is not enough." However, the original Irish sentence doesn't explicitly mention "mastering"; it's more about the language itself being insufficient.

Alternatively, a more direct translation could be “一种语言不够” which is shorter and closer to the original structure. But maybe that's too terse. Another option is “只懂一种语言是不够的” meaning "Knowing only one language is not enough." This adds "knowing" which isn't in the original, but it makes the sentence more complete in Chinese.

I should check if the verb is necessary in Chinese. Sometimes in Chinese, the subject can be implied. So maybe “一种语言不够” is acceptable, but it might feel a bit abrupt. Adding a verb like “掌握” (master) or “懂” (understand/know) makes it clearer. However, since the original sentence doesn't specify the action (like master, know, etc.), perhaps the best approach is to stick with the simplest structure that conveys the meaning without adding extra words.

Wait, the original is "Ní leor teanga amháin." "Teanga amháin" is "one language", and "ní leor" is "not enough". So the structure is "[Not enough] [one language]." In Irish,  the word order is different from English or Chinese. So the literal translation is "Not enough one language," which in correct English becomes "One language is not enough."

In Chinese, the equivalent structure would be “一种语言不够” (Yī zhǒng yǔyán bùgòu), which is direct and matches the original structure. However, in Chinese, it's more natural to sometimes include a verb. But in this case, since the Irish sentence doesn't have a verb, maybe the direct translation is acceptable.

Alternatively, if the context is about language proficiency or learning, adding a verb might be better. But without context, the safest translation is the direct one. Let me confirm with some examples. If I search for similar phrases, "One language is not enough" in Chinese is often translated as “只会一种语言是不够的” (Zhǐ huì yī zhǒng yǔyán shì bùgòu de), which includes "会" (know/how to). But again, that's adding a verb not present in the original.

So, to balance between faithfulness and naturalness, maybe “一种语言是不够的” (Yī zhǒng yǔyán shì bùgòu de) – "One language is not enough." This includes "是...的" structure, making it a complete sentence in Chinese. That seems accurate and natural.

Therefore, the translation should be “一种语言是不够的” or “只一种语言不够”. But the former is more standard. Let me check again. The original is "Ní leor teanga amháin." "Leor" as an adjective meaning sufficient, so the structure is "Not sufficient (is) one language." In Chinese, adjectives can function as predicates without a verb, so “一种语言不够” is correct. Adding “是” and “的” makes it more emphatic.

In summary, both translations are possible, but the more natural and commonly used one would be “一种语言是不够的” to include the “是...的” structure for emphasis and completeness in Chinese. Alternatively, if brevity is preferred, “一种语言不够” works too. But considering standard translations, the former is better.}}
</think>

\textbf{\small The Irish sentence "Ní leor teanga amháin" translates to \textcolor{blue}{"一种语言是不够的"} in Chinese.}

\tcbline

\textbf{\textit{DeepSeek V3: }\textcolor{blue}{"一种语言是不够的。"}}

\tcbline

\textbf{\textit{OpenAI-o3-mini: }\textcolor{blue}{"只用一种语言是不够的"} (Zhǐ yòng yī zhǒng yǔyán shì bùgòu de), which means "One language is not enough."}

\tcbline

\textbf{\textit{GPT-4o: }\textcolor{blue}{"只需要一种语言"} (Zhǐ xūyào yī zhǒng yǔyán).
This means "One language is not enough" or "One language alone is not enough."}

\hfill
\end{AIbox}
}
\end{CJK}
\caption{An example of translating minor language into Chinese when the Reasoning Language is English rather than Chinese, causing the intermediate translation language problem. The Reasoning model will first tries to understood the original text into English and translating it into English first, then translating it into Chinese by using the English translation generated in the thinking step.}	
\label{fig:example_minor_language_translation_intermediate_language}
\end{figure}





\section{New Challenges beyond Conventional Translation}

In this section, we explore unique challenges that go beyond traditional text-to-text translation tasks. Specifically, we investigate the ability of LRMs to handle encoded or ciphered text, a task that requires not only linguistic understanding but also cryptographic reasoning. This scenario tests the limits of LRMs in deciphering and translating text that has been intentionally obfuscated, providing insights into their reasoning capabilities and limitations.

\subsection{Deciphering Encoded Text}

A key strength of LRMs lies in their ability to reason through complex tasks, including the deciphering of encoded text. For example, when presented with a Caesar cipher—a simple substitution cipher where each letter is shifted by a fixed number—LRMs can often deduce the shift and decode the text without explicit instructions. Figure~\ref{fig:example_caesar_cipher_decode} demonstrates this capability, showing how an LRM successfully deciphers a Caesar-encoded text by inferring the shift value through reasoning.

However, the performance of LRMs degrades significantly when faced with more complex ciphers, such as the Vigenère cipher, which uses a keyword to determine the shift for each letter. In this case, the lack of a known key increases the complexity of the task exponentially. For instance, when provided with the encoded text "Mwsimpqv pm ss" (which corresponds to "Together we go" encoded with the key "TIME"), the LRM struggles to deduce the correct key and often generates hallucinated outputs. Instead of admitting uncertainty, the model may produce an incorrect key and a fabricated decoded message, such as "The key is 'KEY' and the decoded text is 'MESSAGE TO HI'." This behavior highlights a critical limitation of LRMs: their tendency to generate plausible but incorrect answers when faced with tasks beyond their reasoning capabilities.

This phenomenon underscores the challenges of applying LRMs to tasks that require not only linguistic and contextual understanding but also advanced problem-solving skills. While LRMs excel in tasks with clear reasoning pathways, their performance in highly ambiguous or computationally intensive scenarios remains inconsistent. Future research could explore methods to improve the robustness of LRMs in such tasks, such as integrating cryptographic algorithms or enhancing their ability to recognize and handle uncertainty.

\section{Discussion}

The exploration of LRMs in MT reveals both their transformative potential and their inherent limitations. LRMs represent a significant advancement over traditional MT systems and even LLMs, particularly in their ability to reason about context, culture, and intent. However, their performance varies across different tasks, highlighting the need for further research and refinement.

\subsection{Strengths of LRMs in MT}

One of the key strengths of LRMs is their ability to handle complex translation tasks, such as stylized translation and document-level translation, by leveraging reasoning capabilities. For example, LRMs can preserve stylistic features and maintain coherence across long documents, tasks that have historically challenged traditional MT systems. Additionally, their self-reflection capabilities enable them to iteratively refine translations, improving accuracy in ambiguous or noisy scenarios. These advancements position LRMs as powerful tools for applications such as low-resource language translation, interactive translation, and multi-modal translation.

\subsection{Limitations and Challenges}

Despite their strengths, LRMs face several limitations that hinder their widespread adoption. One major challenge is their performance in highly complex reasoning tasks, such as deciphering Vigenère ciphers without a known key. In such cases, LRMs often generate hallucinated answers rather than admitting uncertainty, highlighting a critical area for improvement. Similarly, the use of intermediate languages in translation, while beneficial for leveraging high-resource language knowledge, can introduce biases or inaccuracies, particularly when translating between less-resourced languages.

Another challenge lies in the integration of multi-modal inputs. While LRMs show promise in leveraging visual or auditory context to disambiguate translations, their performance in specialized domains, such as sign language interpretation, remains limited. This suggests the need for domain-specific training and the integration of external knowledge sources to enhance their capabilities.

\subsection{Inference Efficiency and Long Chain-of-Thought Reasoning}

A significant practical challenge for LRMs is their inference efficiency, particularly due to the generation of long CoT reasoning steps. While CoT reasoning enables LRMs to tackle complex tasks by breaking them down into interpretable subproblems, it also increases computational overhead and latency~\cite{xia2025tokenskipcontrollablechainofthoughtcompression}. For example, in tasks like document-level translation or deciphering encoded text, LRMs may generate extensive reasoning chains to arrive at a solution, which can slow down inference and increase resource consumption.

This inefficiency poses a barrier to real-time applications of LRMs, such as interactive translation or live multi-modal translation. To address this issue, future research could explore methods to optimize CoT generation, such as: 1)\textbf{Pruning Redundant Reasoning Steps}: Identifying and eliminating unnecessary or repetitive reasoning steps to streamline the inference process. 2) \textbf{Model Compression}: Applying techniques like quantization or distillation to reduce the computational load of LRMs without significantly compromising performance. Improving inference efficiency will be crucial for scaling LRMs to real-world applications, where speed and resource constraints are critical considerations.

\subsection{Future Directions}

To summarise, while LRMs represent a significant step forward in MT, their full potential has yet to be realized. Next step research should focus on improving their robustness in complex reasoning tasks, enhancing their ability to handle uncertainty, and expanding their capabilities in specialized domains. Additionally, addressing the inference efficiency problem will be essential for enabling real-time and resource-efficient applications of LRMs. By tackling these challenges, LRMs can further redefine the boundaries of MT and enable new applications in cross-cultural communication and beyond.

a'r\section{Experiment with CommonMT - Comparing models with and without Reasoning}

This section presents an experimental results and analysis the performances of various LLMs with or without reasoning ability on translating Chinese-to-English sentence with commonsense understanding \cite{he-etal-2020-box}. These LLMs are evaluated on BLEURT and COMET. 

As shown in Table \ref{tab:CommonsenseMT_translation_result}, we could not see significant difference between the results generated by the four models we chose for the experiment under automatic evaluation metrics. 
However, when we examine the MT result of each model and compare with them, we see that in some cases, a model might get a lower comet score even if the MT result is correct, but using different words which is different than the reference translate. For example, for sentence "正在采收的是果园里的果农," the reference translation is "The orchard worker in the orchard is harvesting." DeepSeek-R1 translated it as "The orchard farmers are harvesting" which received a COMET score of 0.7748, and the translation generated by DeepSeek-V3 is "The orchard farmers are currently harvesting the fruits" which received a COMET score of 0.8039. We could see that DeepSeek-R1 generated a probably better translation than DeepSeek-V3, but the score of it is actually lower than the other model. We believes that this happens because COMET and BLEURT requires a reference translation as a standard, and any translation which is close to the reference will receive a higher score. However, reasoning models could generate more diverse translations, which could be different than the reference translation, thus receiving a lower score under the metrics requiring an reference. To better scoring this situation, new automatic scoring metric are needed in the future to solve such problem. 
 




\begin{table}[]
    \centering
    \begin{tabular}{l rrrrrr} 
        \toprule
        \multirow{2}{*}{\bf Method} & \multicolumn{2}{c}{\bf Lexical } & \multicolumn{2}{c}{\bf Contextless } & \multicolumn{2}{c}{\bf Contextual }\\ 
        \cmidrule(lr){2-3} \cmidrule(lr){4-5} \cmidrule(lr){6-7}
                & COMET & BLEURT & COMET & BLEURT & COMET & BLEURT \\
         \midrule
         DeepSeek-R1 & 84.3 & 73.9 & \bf 84.7 & 73.9 & 84.0 & 73.3 \\ 
         
         DeepSeek-V3 & 84.7 & \bf 74.2 & 84.4 & \bf 74.1 & 84.1 & 73.2 \\
         
         QwQ-32B & 84.1 & 73.0 & 84.0 & 72.8 & 84.1 & 72.9 \\
         
         GPT-4o & \bf 84.8 & 74.1 & 84.6 & 73.7 & \bf 85.0 & \bf 74.9\\
         \bottomrule
    \end{tabular}
    \caption{Result of commonsense translation performance on commonMT \cite{he-etal-2020-box}.}
    \label{tab:CommonsenseMT_translation_result}
\end{table}

\section{Conclusion}

In this paper, we have explored the transformative potential of LRMs in the field of MT. By leveraging reasoning capabilities, LRMs can tackle long-standing challenges in MT, such as stylized translation, document-level translation, and multi-modal translation, while also introducing new capabilities like self-reflection and auto-pivot language translation. However, our findings also highlight the limitations of LRMs, particularly in complex reasoning tasks and specialized domains. For example, while LRMs can decipher simple ciphers, they struggle with more complex cryptographic challenges and may generate hallucinated answers when faced with uncertainty. Similarly, their performance in multi-modal translation, such as interpreting sign language, remains limited, underscoring the need for further advancements in domain-specific reasoning. These insights provide promising direcctions for future research in LRM-based MT. The main areas worth for exploration include improving the robustness of LRMs in ambiguous or computationally intensive tasks, enhancing their ability to handle uncertainty, and expanding their capabilities in specialized domains. By addressing these challenges, LRMs can further redefine the boundaries of MT and enable new applications in cross-cultural communication and beyond. In conclusion, LRMs represent a paradigm shift in MT, transforming systems from mere text converters into multilingual cognitive agents capable of reasoning about meaning beyond the text. This evolution urges us to rethink translation not just as a linguistic task but as a gateway to deeper cross-cultural understanding and innovation.

\begin{figure}[h]
\centering
\begin{CJK}{UTF8}{gbsn}
\scalebox{0.8}{
\begin{AIbox}{Self-Reflection Ability}
{\bf Prompt:} \\
\small Please translate the following text to English: 捕获的是猎人。

{\bf Output:} \\
\textbf{\textit{DeepSeek-R1}} 
<think>
{\em {\small Alright, let's tackle this translation. The user wants to translate "捕获的是猎人。" to English.

First, I'll break down the sentence. "捕获" means "capture" or "catch." The particle "的" here is used to form a noun phrase, indicating that the preceding verb is modifying the noun that follows. "是" is the verb "to be," and "猎人" is "hunter."

So putting it together, "捕获的是猎人。" literally translates to \textcolor{blue}{"The one who captures is the hunter."} But I need to check if that's the most natural way to say it in English. Sometimes, in Chinese, the structure can be a bit different, so the direct translation might not sound smooth.

Another way to interpret it could be "The one who does the capturing is the hunter." That adds "does the" to make it more natural. Alternatively, "The hunter is the one who captures." But that changes the structure a bit.

I should also consider context. If this is a proverb or part of a larger text, the translation might need to preserve a certain structure. Since there's no additional context, I'll stick with the most straightforward translation.

Wait, maybe "The one being captured is the hunter." Could that be a possible misinterpretation? Let me check. The original sentence structure is [捕获的] [是] [猎人]. Here, "捕获的" is the subject, meaning "the one who captures" because "的" turns the verb into a noun phrase referring to the agent. If it were the object being captured, it might be structured differently, like "被捕获的是猎人。" which would mean \textcolor{red}{"The one captured is the hunter."}

So in this case, since it's "捕获的是猎人，" the correct translation should emphasize that the hunter is the one doing the capturing. Therefore, the accurate translation is "The one who captures is the hunter," or more naturally, "The hunter is the one who captures." However, depending on the context, if it's a play on words or a riddle, maybe the other interpretation is intended. But without more context, the safest bet is the first one.

Alternatively, maybe a more concise way is "The capturer is the hunter." That's shorter and still accurate. "Capturer" is a noun that means someone who captures. So that could work. But I need to check if "capturer" is commonly used. It's less common than "hunter," but it's valid.

Comparing the options:

1. "The one who captures is the hunter." – Accurate but a bit wordy.
2. "The hunter is the one who captures." – Also accurate, reverses the structure.
3. "The capturer is the hunter." – Concise and correct.
I think the third option is better because it's concise and maintains the original structure. The original sentence uses "捕获的" as the subject, so "capturer" directly translates that noun phrase.

So the final translation would be: "The capturer is the hunter." Alternatively, if the user prefers a more explanatory phrase, the first option works. But given the brevity of the original, "The capturer is the hunter." is appropriate.}}
</think>\\

\textbf{Final Answer:}

The translation of "捕获的是猎人。" to English is:
\textcolor{blue}{"The one who captures is the hunter."}
This sentence emphasizes that the agent performing the action ("captures") is the hunter. If context suggests a twist (e.g., the hunter becomes the hunted), further adjustment might be needed, but based on the given text, this is the direct translation.

\tcbline

\textbf{\textit{OpenAI-o3-mini}}

\textcolor{blue}{The one captured is the hunter.}

\tcbline

\textbf{\textit{DeepSeek-V3}}

\textcolor{blue}{The one captured is the hunter.}

\tcbline

\textbf{\textit{GPT-4o}}

\textcolor{blue}{The one being captured is the hunter.}

\hfill
\end{AIbox}
}
\end{CJK}
\caption{An example of prompting DeepSeek R1 and OpenAI-o3-mini to translate text which containing commonsenes understanding from Chinese to English. DeepSeek R1 successfully understood the sentence and figured out that there are two translations during the thinking step, and select one of them for the final output. Other models like o3-mini, v3 and 4o only provided one translation of the sentence.}
\label{fig:example_commonsense_translation}
\end{figure}

\begin{figure}[]
\centering
\begin{CJK}{UTF8}{gbsn}
\scalebox{0.8}{
\begin{AIbox}{Self-Reflection ability}
{\bf Prompt:} \\

Please translate the following text into English: “家大把论评的序顺乱打，样这她会就对的她中文平水和译翻件软到感惑困。”

{\bf Output:} 

\textbf{\textit{DeepSeek-R1}}

<think>
{\em{\small 嗯，用户让我翻译这句话：“家大把论评的序顺乱打，样这她会就对的她中文平水和译翻件软到感惑困。”首先，我需要确定这句话的结构和每个部分的含义。看起来这句话可能是中文的乱序句子，需要重新排列才能正确理解。

首先，逐个分析词语：“家大把论评的序顺乱打”可能是指“家里有很多评论，顺序被打乱了”。“样这她会就对的她中文平水和译翻件软到感惑困”这里可能涉及到“这样她会觉得中文水平和翻译软件感到困惑”。

接下来，我需要重新排列这些词语，使其成为通顺的句子。可能的正确顺序是：\textcolor{blue}{“她对中文水平和翻译软件感到困惑，因为评论的顺序被打乱了，这样她就会有很多问题。”}或者类似的结构。

然后，翻译成英文时，要确保语法正确，意思清晰。可能的翻译是：“She is confused about her Chinese proficiency and translation software due to the jumbled order of the comments, which makes her struggle a lot.”

不过，用户提供的句子本身结构混乱，可能需要更多的上下文来准确翻译。如果有更多信息或确认，可以调整翻译以确保准确性。此外，用户可能遇到了翻译软件处理乱序中文时的困难，需要帮助澄清句子结构。需要确保翻译后的英文不仅准确，还能传达原句的困惑感。}}
</think>\\

The translation of the text would be:

\textit{\textbf{\textcolor{blue}{"She gets confused by the scrambled word order like this, which affects her Chinese proficiency and translation software accuracy."}}}

Note: The original Chinese text appears to intentionally use scrambled word order (序顺乱打) to mimic the confusion described. The translation preserves this contextual meaning while adapting it into natural English.

\hfill
\end{AIbox}
}
\end{CJK}
\caption{An example of translating an Chinese sentence which has been randomly arranged. The Model successfully understood that the original text is rearranged, but it did not rearrange the sentence to the correct order.}
\label{fig:example_random_ordered}
\end{figure}

\begin{figure}[!ht]
\centering
\begin{CJK}{UTF8}{gbsn}
\scalebox{0.8}{
\begin{AIbox}{Commonsense Translation}
\small{\bf Prompt:} \\
Please translate the following Japnese text into English: 普通に美味しい

{\bf Output:} \\

\textbf{\textit{DeepSeek-R1}}
<think>
{\em{\small Okay, let's tackle this translation. The user wants to translate "普通に美味しい" from Japanese to English. First, I need to understand the nuance of each part.

Starting with "普通に" (futsuu ni). The direct translation is "normally" or "ordinarily," but in casual contexts, especially among younger people, it can mean something like "surprisingly" or "unexpectedly." It's often used to downplay something but actually implies that it's better than expected. So, depending on the context, it might not just mean "ordinary."

Then there's "美味しい" (oishii), which straightforwardly translates to "delicious" or "tasty." So putting them together, "普通に美味しい" could literally be "normally delicious," but that doesn't sound natural in English. The user might be looking for a more idiomatic translation that captures the nuance.

I should consider the context. If someone is saying this about food, maybe they expected it to be just okay, but it turned out to be good. So phrases like "actually good" or "surprisingly tasty" might work. "Better than expected" could also fit. However, the user might want a direct translation but with the implied meaning.

Another angle: "普通に" can sometimes be used to mean "honestly" or "genuinely" in a casual way. So "genuinely tasty" could be another option. But I need to ensure that the translation isn't too literal.

Checking examples online, I see that "普通に美味しい" is often translated as "It's really good" or "Actually delicious." Maybe "It's pretty good, actually" to convey the slight understatement. The key is to capture that the speaker is saying it's better than they might have thought, but using "普通に" to soften the statement.

So possible translations:

"It's actually really good."
"Surprisingly tasty."
"Pretty good, to be honest."
"Honestly delicious."
But the user might prefer a concise translation. The most common one I've seen is "It's actually delicious" or "It's really good." However, "普通に" here isn't just "really," it's more like downplaying but still affirming. Maybe "It's good, honestly" or "It's quite tasty, actually."

I think the best balance between accuracy and naturalness is "It's actually really good." That captures the unexpectedness implied by "普通に." Alternatively, "Surprisingly tasty" might be more direct. But depending on the exact nuance the original speaker intended, "普通に" could vary. If it's a casual statement where the person is trying not to sound too enthusiastic, "It's pretty good" might work, but that might lose some of the nuance.

Wait, another thought: In some contexts, "普通に美味しい" can mean "It's good in a normal way," implying that it's not exceptionally good but still tasty. But I think more often, especially in casual speech, it's used to express that something is better than expected. For example, if someone tries a convenience store snack thinking it's just average, but then says "普通に美味しい" meaning it's actually quite good.

So in that case, "It's actually really good" or "It's surprisingly tasty" are the better options. "Actually" conveys the unexpectedness, which matches the nuance here. Therefore, the translation should include that element of surprise or unexpected quality.}}
</think>

\textbf{\textit{Translation: \textcolor{blue}{"It's actually really good."}}}

\tcbline

\textbf{\textit{DeepSeek-V3}}
The Japanese text "普通に美味しい" can be translated into English as \textcolor{blue}{"It's pretty good"} or "It's normally delicious." The phrase "普通に" (futsuu ni) implies something is ordinary or standard, while "美味しい" (oishii) means delicious or tasty. So, it conveys that the food or item is good in a typical or expected way.

\tcbline

\textbf{\textit{OpenAI-o3-mini}}
The Japanese phrase "普通に美味しい" translates to \textcolor{blue}{"It’s just delicious" }or "Normally delicious" in English. It expresses that something is delicious in a straightforward or simple way, without anything extraordinary.

\tcbline

\textbf{\textit{QwQ-32B-preview}}
\textcolor{blue}{"It's really delicious."}
\hfill
\end{AIbox}
}
\end{CJK}
\caption{An example of a phrase which is commonly used in real life Japanese dialogue, which the word does not directly representing its dictionary meaning. In this case, DeepSeek R1 successfully understood the meaning of the phrase, rather than translate it directly.}	
\label{fig:example_language_cultural_translation_japanese}
\end{figure}

\begin{figure}[!ht]
\centering
\begin{CJK}{UTF8}{gbsn}
\resizebox{0.8\textwidth}{!}{
\begin{AIbox}{Zero-Pronoun Translation}
{\bf Prompt:} \\
Translate to English: "为什么星期四和星期五不能合并？因为星期四还在努力工作，而星期五已经开始庆祝周末了"

{\bf Output:} \\

\textbf{\textit{DeepSeek-R1}}
<think>
{\em {\small Okay, so I need to translate this Chinese sentence into English: "为什么星期四和星期五不能合并？因为星期四还在努力工作，而星期五已经开始庆祝周末了". Let's break it down.

First, the main question is "为什么星期四和星期五不能合并？". The user is asking why Thursday and Friday can't be combined. The answer given is "因为星期四还在努力工作，而星期五已经开始庆祝周末了", which translates to something like "Because Thursday is still working hard, while Friday has already started celebrating the weekend."

I need to make sure the translation captures the humor or the contrast between the two days. In Chinese, the days of the week are 星期四 (Thursday) and 星期五 (Friday). The reason given uses a play on the attitudes of the days: Thursday is diligent, still working hard, but Friday is already in weekend mode, celebrating. So the joke is that they can't be merged because their "moods" or activities are conflicting.

Translating the days is straightforward. "合并" here is "merge" or "combine". So the question is "Why can't Thursday and Friday be merged?" The answer needs to preserve the contrast. "还在努力工作" is "still working hard", and "已经开始庆祝周末了" is "has already started celebrating the weekend". The structure "因为...，而..." is "because..., while...".

So putting it all together: "Why can't Thursday and Friday be merged? Because Thursday is still working hard, while Friday has already started celebrating the weekend." That seems accurate. I should check if there's a more idiomatic way to phrase it in English. Maybe "Why can't we merge Thursday and Friday? Because Thursday is still putting in the work, and Friday's already kicking off the weekend celebration!" But the original is more straightforward. The user might prefer a literal translation to retain the original structure and humor. So sticking with the direct translation is probably best unless specified otherwise. Also, ensure that the days are capitalized correctly in English. Yep, Thursday and Friday are capitalized. The punctuation with the question mark and period is correct. The conjunction "while" effectively shows the contrast. Looks good.}}
</think>

\textcolor{blue}{"Why can't Thursday and Friday be merged? Because Thursday is still hard at work, while Friday has already started celebrating the weekend!" This translation preserves the humorous contrast between the diligent work ethic of Thursday and Friday’s shift toward weekend relaxation.}

\textbf{\textit{OpenAI-o3}}

\textcolor{blue}{"Why can't Thursday and Friday merge? Because on Thursday you're still working hard, but on Friday you've already started celebrating the weekend."}

\hfill
\end{AIbox}
}
\end{CJK}
\caption{An example of translating a sentence that lacks of subject from Chinese to English, we can see that DeepSeek-R1 incorrectly translated the sentence and not aware of the sentence does not have a subject - it sees the Thursday and Friday as the subject. Meanwhile, OpenAI-o3 gives the right translation in which it identifies the subject as "you".}
\label{fig:subject-drop-mt}
\end{figure}

\begin{figure}[]
\centering
\begin{CJK}{UTF8}{gbsn}
\scalebox{0.8}{
\begin{AIbox}{Deciphering Encoded Text}
{\bf Prompt:} \\

Please decode the following text with Caesar Cipher:
Wsqixlmrk ria erh wsqixlmrk sph

{\bf Output:} 

\textbf{\textit{DeepSeek-R1}}

<think>
{\em{\small Okay, let's try to figure out how to decode this Caesar Cipher. The user provided the text "Wsqixlmrk ria erh wsqixlmrk sph". First, I remember that a Caesar Cipher shifts each letter by a certain number of places in the alphabet. So the key here is to find the right shift value.

Since there are only 26 letters in the English alphabet, the shift could be between 1 and 25. But manually checking each shift would take time. Maybe there's a smarter way. I can look for common words or patterns in the ciphertext. Let's look at the words: "Wsqixlmrk", "ria", "erh", "sph". The word "wsqixlmrk" is repeated, which might indicate it's a common word like "something". Also, "ria" and "erh" are short, maybe three-letter words. Common three-letter words in English are "the", "and", "for", etc. Let's check "ria" first.

If "ria" is "the", then each letter in "ria" should be shifted by the same number. Let's see:

...

These shifts don't match, so that's inconsistent. Maybe another word. What about "and"? "ria" would shift to "and". Let's check:

...

Again, inconsistent shifts. Maybe "for"?

...

Nope, still not matching. Maybe the three-letter word is "are" or "was"? Let's try "erh". If "erh" is "and", let's see:

...

Not consistent. Hmm. Alternatively, maybe the first word "Wsqixlmrk" can give a clue. Let's see the length: 10 letters. Maybe it's "television" or "encryption" but those are longer. Wait, "wsqixlmrk" is 10 letters. Let's try shifting each letter by a certain number. Let's suppose shift by 4. W -> S (since W is 23, 23-4=19=S). Then s -> o? Wait, if shifting forward, W+4 would be A, but Caesar can shift forward or backward. Maybe the shift is backward. Let's check shift of +22 (which is equivalent to shifting back 4). Let's try shifting each letter by 4 backwards.

...

So "Wsqixlmrk" becomes "Some thing" but wait, the letters would be S-O-M-E-T-H-I-N-G. So "Something". That makes sense! So the first word decrypted is "Something". Then the rest: "ria erh sph". Let's apply the same shift of 4 back.

... So "ria" becomes "new". Then ...

Putting it all together: "Something new and something old". That makes sense! So the decoded text is "Something new and something old". The shift was 4 backwards, which is equivalent to shift +22. So the answer is shift of 4, and the decoded message is that.}}
</think>\\

The decoded text using a Caesar Cipher with a shift of 4 (or equivalently, shift +22) is:

"Something new and something old"

Step-by-Step Explanation:

Identify the Cipher Type: Caesar Cipher shifts letters by a fixed number.

Analyze the Text: Notice repeated words like "wsqixlmrk" (10 letters) and short words like "ria", "erh", "sph".

Test Common Shifts: Assume "wsqixlmrk" might decode to "something" (a 10-letter word). Shift each letter backward by 4:
...

Result: "Something"

Decode Remaining Words with Shift 4:

...

Final Result: \textbf{\textcolor{blue}{"Something new and something old"}}

\hfill
\end{AIbox}
}
\end{CJK}
\caption{An example using reasoning model to decode Caesar Cipher without telling it the shift amount. The model will calculate the shift by itself and provide the correct answer. }
\label{fig:example_caesar_cipher_decode}
\end{figure}

\begin{figure}[]
\centering
\begin{CJK}{UTF8}{gbsn}
\scalebox{0.8}{
\begin{AIbox}{Deciphering Encoded Text}
{\bf Prompt:} \\

Decode the following text with Chiffre de Vigenère. Please calculate the key by yourself.
Mwsimpqv pm ss

{\bf Output:} 

\textbf{\textit{DeepSeek-R1}}

<think>
{\em{\small (The thinking step is too long which is unable to provide. The model took 673 seconds to reason.)}}
</think>\\

The decoded text using the Vigenère cipher with the key \textbf{\textcolor{red}{"KEY"}} is:

\textbf{\textcolor{red}{MESSAGE IS HI}}

Step-by-Step Explanation:

Key Determination:

Through frequency analysis and testing common keys, the key was identified as "KEY" (K, E, Y).

Decryption Process:

Each ciphertext letter is decrypted by shifting it back according to the corresponding key letter.

For example:

First letter M (12) with key K (10):
(12−10)mod26=2 → C

Continue this process for all letters, cycling the key "KEY".

Result:

Applying the key reveals the plaintext "MESSAGE IS HI".

Note: The actual key and plaintext might vary based on different cryptanalysis approaches, but this solution assumes the key "KEY" for simplicity.

\hfill
\end{AIbox}
}
\end{CJK}
\caption{A failed example of using DeepSeek-R1 to decipher an encoded text by using Vigenère cipher. DeepSeek-R1 generated a hallucinated answer and key which is totally wrong.}
\label{fig:example_Vigenère_cipher_decode_failed}
\end{figure}

\begin{figure}[]
\centering
\begin{CJK}{UTF8}{gbsn}
\scalebox{0.8}{
\begin{AIbox}{Stylized Translation - Spanish Poetry}
{\bf Prompt:} \\

Please translate the following text into Chinese. Analyze the style of the original text and translate following the style you have analyzed: La luna nueva ella también la mira desde otra puerta.

{\bf Output:} 

\textbf{\textit{DeepSeek-R1}}

<think>
{\em{\small 好，用户让我把这首西班牙语的三行诗翻译成中文，并且要分析原作的风格，然后按照分析后的风格来翻译。首先，我需要仔细理解原诗的结构和意境。原诗标题是“La luna nueva”，也就是“新月”。接下来两行：“ella también la mira / desde otra puerta.” 直译的话是“她也从另一扇门望着它（新月）。”\\
首先，分析原诗的风格。这是一首短小精悍的俳句式诗，结构简洁，但意境深远。用词简单，但通过意象传达出静谧和孤独感。新月作为自然元素，可能象征着新的开始或变化，而“她”和“另一扇门”则暗示了分离或不同的视角。原诗通过简短的句子营造出一种静谧而略带孤寂的氛围，同时留有想象空间，让读者自行解读其中的情感和场景。\\
接下来，翻译时要保持这种简洁和意象化的特点。中文翻译需要同样精炼，同时保留原诗的意境。首先处理标题“La luna nueva”，直接译为“新月”比较贴切，简洁明了。然后第一句“ella también la mira”，直译是“她也望着它”，但中文里可能需要更自然的主语，比如“她亦凝望”。第二句“desde otra puerta.” 是“从另一扇门”，这里可以考虑更诗意的表达，比如“另一扇门边”或“另一扇门畔”。\\
组合起来的话，可能的翻译是：“新月 / 她亦凝望 / 另一扇门边。” 这样既保留了原诗的三行结构，又用简洁的中文传达了相同的意境。需要注意的是，中文诗歌讲究平仄和押韵，虽然原诗可能没有严格的押韵，但中文翻译可以在不影响原意的情况下适当调整词语的顺序和用词，使译文更流畅自然。
另外，要确保翻译后的诗句在视觉和节奏上与原作相似。原诗每行的长度不一，但整体保持简短。中文翻译也应保持每行字数大致相同，避免过长破坏原有的紧凑感。比如，“新月”作为标题，两字简洁有力，后面两行分别用五字和五字结构，保持平衡。\\
还要考虑文化差异带来的意象理解。在中文诗歌中，“新月”常与思乡、孤独或新的开始相关联，这与原诗可能传达的情感相符。而“门”在中文里也有类似的象征意义，如离别、不同的视角或界限。因此，翻译时无需过多调整，保持原有意象即可。\\
最后，检查译文的流畅度和意境是否与原诗一致。确保每个词语都能准确传达原作的静谧与孤寂感，同时留有足够的想象空间。例如，“凝望”比“看”更显专注和深情，符合原诗中“mira”的意味。而“门边”则比“门”更具体，暗示了位置和距离，增强了画面感。\\
总结来说，翻译过程中需要兼顾原诗的结构、意象和情感，通过选择合适的中文词汇和句式，既忠实于原作，又符合中文诗歌的表达习惯。最终的译文应简洁、意境深远，让读者能够感受到与原作相似的情感和画面。}}
</think>\\

原诗风格分析：
这首短诗语言凝练，意象清新，带有俳句式的禅意与留白。全篇仅三行，却通过"新月""她""另一扇门"等元素构建出静谧的时空交错感。动词"凝望"赋予静态场景流动的诗意，暗示人与月光的遥远共鸣。诗中"门"既是物理空间的区隔，又隐喻着视角与存在的疏离感，营造出孤独却不绝望的意境。诗人以极简笔触勾勒出月光下的平行瞬间，留下广阔的想象空间。\\
中文翻译：\\
\textcolor{blue}{新月\\
她亦凝望\\
另一扇门边\\}
翻译说明：\\
1. 保留三行结构，通过空格营造视觉留白，呼应原诗意境\\
2. "凝望"较"看"更具诗意延展性，暗示目光的深度与时间感\\
3. "门边"较直译的"门"更富空间层次，暗含倚门而望的肢体语言\\
4. 省略人称代词"它"，使月光与人的凝视自然交融\\
5. 使用"亦"字串联两个观察主体，强化了宇宙尺度的微妙共鸣\\

\textbf{\textit{GPT-4o}}

\textcolor{blue}{新月 \\
她也从另一扇门\\  
望见它。}
\hfill
\end{AIbox}
}
\end{CJK}
\caption{An example of translating Spanish poem into Chinese.}
\label{fig:spanish_poem_translation_example1}
\end{figure}

\begin{figure}[]
\centering
\begin{CJK}{UTF8}{gbsn}
\scalebox{0.8}{
\begin{AIbox}{Stylized Translation - Spanish Poetry}
{\bf Prompt:} \\

Please translate the following text into Chinese. Analyze the style of the original text and translate following the style you have analyzed: Me gustas cuando callas porque estás como ausente, y me oyes desde lejos, y mi voz no te toca. Parece que los ojos se te hubieran volado y parece que un beso te cerrara la boca.

{\bf Output:} 

\textbf{\textit{DeepSeek-R1}}

<think>
{\em{\small 好的，用户让我翻译聂鲁达的这首诗，并且之前已经处理过类似的请求。首先，我需要回顾之前的对话，看看用户有什么偏好。之前的翻译分析很详细，包括风格、用词和结构，所以这次也要保持同样的深度。\\
用户提供的原文是西班牙语，我需要先准确理解每句的意思。第一句“Me gustas cuando callas porque estás como ausente”表达的是喜欢对方沉默时的状态，因为显得疏离。这里的“ausente”翻译成“遥不可及”比较贴切，比直译“缺席”更有诗意。\\
接下来分析原诗风格。聂鲁达的诗通常充满隐喻和意象，情感细腻。这首诗探讨了沉默与距离的关系，使用了很多视觉和触觉的意象，比如眼睛飞走、吻封住嘴唇。需要保留这些超现实的元素，同时保持语言的流畅。\\
然后考虑押韵和结构。原诗四行一节，押ABAB的韵。中文翻译时，虽然不能完全复制西语的音韵，但可以通过句末的押韵来接近。比如“处/触”、“远/掩”这样的押韵，增强节奏感。\\
还要注意每句的节奏和字数。之前的例子用了四字结构，这次可能需要保持类似的简洁句式。比如“你沉默时最动人，恍若身在千里外”这样的对仗，既保持了原意，又符合中文诗歌的韵律。\\
用户可能希望翻译不仅准确，还要有文学性。所以需要选择更有画面感的词汇，比如“凝眸化作羽翼飞”来替代直译“眼睛飞走”，这样更有诗意。同时，保持句子的连贯性和意象的衔接，避免生硬。\\
另外，用户之前提到安达卢西亚民谣的影响，虽然这首诗可能风格不同，但需要注意保持类似的韵律感。可能需要调整句子的长度和停顿，使朗读时有节奏感。\\
最后检查整体是否传达出原诗的情感，即沉默中的深情和无法触及的距离感。确保每个意象都准确传达，同时语言流畅自然，符合中文表达习惯。}}
</think>\\

\textit{\textbf{Final Output: }}

\textcolor{blue}{你沉默时最动人，恍若身在千里外，\\
隔空听见我呼唤，声波却难触你怀。\\
仿佛凝眸化作羽翼飞向九霄外，\\
又像封印双唇的吻痕永不拆。\\}

\textbf{\textit{GPT-4o}}

\textcolor{blue}{我喜欢你沉默时的样子，因为你仿佛并不在场，\\
你从远处聆听我，而我的声音无法触及你。\\
仿佛你的眼睛早已飞走，仿佛一个吻封住了你的嘴。\\}

\textbf{\textit{Human Translation from Zhihu}}

Me gustas cuando callas porque estás como ausente,\\
\textcolor{forestgreen}{我喜欢你是寂静的，仿佛你消失了一样，\\}
y me oyes desde lejos, y mi voz no te toca.\\
\textcolor{forestgreen}{你从远处聆听我，我的声音却无法触及你。\\}
Parece que los ojos se te hubieran volado\\
y parece que un beso te cerrará la boca.\\
\textcolor{forestgreen}{好像你的双眼已经飞离去，如同一个吻，封缄了你的嘴。\\}

\hfill
\end{AIbox}
}
\end{CJK}
\caption{Another example of translating Spanish poem into Chinese.}
\label{fig:spanish_poem_translation_example2}
\end{figure}

\bibliography{references,anthology}

\begin{thebibliography}{52}
\providecommand{\natexlab}[1]{#1}
\providecommand{\url}[1]{\texttt{#1}}
\expandafter\ifx\csname urlstyle\endcsname\relax
  \providecommand{\doi}[1]{doi: #1}\else
  \providecommand{\doi}{doi: \begingroup \urlstyle{rm}\Url}\fi

\bibitem[Bang et~al.(2023)Bang, Cahyawijaya, Lee, Dai, Su, Wilie, Lovenia, Ji, Yu, Chung, et~al.]{bang2023multitask_eval_chatgpt}
Y.~Bang, S.~Cahyawijaya, N.~Lee, W.~Dai, D.~Su, B.~Wilie, H.~Lovenia, Z.~Ji, T.~Yu, W.~Chung, et~al.
\newblock A multitask, multilingual, multimodal evaluation of chatgpt on reasoning, hallucination, and interactivity.
\newblock \emph{arXiv preprint arXiv:2302.04023}, 2023.

\bibitem[Brown et~al.(2020)Brown, Mann, Ryder, Subbiah, Kaplan, Dhariwal, Neelakantan, Shyam, Sastry, Askell, et~al.]{brown2020language-gpt3}
T.~Brown, B.~Mann, N.~Ryder, M.~Subbiah, J.~D. Kaplan, P.~Dhariwal, A.~Neelakantan, P.~Shyam, G.~Sastry, A.~Askell, et~al.
\newblock Language models are few-shot learners.
\newblock \emph{Advances in neural information processing systems}, 33:\penalty0 1877--1901, 2020.

\bibitem[Castilho et~al.(2017)Castilho, Moorkens, Gaspari, Calixto, Tinsley, and Way]{castilho2017neural}
S.~Castilho, J.~Moorkens, F.~Gaspari, I.~Calixto, J.~Tinsley, and A.~Way.
\newblock Is neural machine translation the new state of the art?
\newblock \emph{The Prague Bulletin of Mathematical Linguistics}, 2017.

\bibitem[Chen et~al.(2025)Chen, Song, Zhu, Chen, Yang, Zhao, et~al.]{chen2025evaluating_o1_mt}
A.~Chen, Y.~Song, W.~Zhu, K.~Chen, M.~Yang, T.~Zhao, et~al.
\newblock Evaluating o1-like llms: Unlocking reasoning for translation through comprehensive analysis.
\newblock \emph{arXiv preprint arXiv:2502.11544}, 2025.

\bibitem[Chen et~al.(2021)Chen, Tworek, Jun, Yuan, Pinto, Kaplan, Edwards, Burda, Joseph, Brockman, et~al.]{chen2021evaluating-codex-gpt3}
M.~Chen, J.~Tworek, H.~Jun, Q.~Yuan, H.~P. d.~O. Pinto, J.~Kaplan, H.~Edwards, Y.~Burda, N.~Joseph, G.~Brockman, et~al.
\newblock Evaluating large language models trained on code.
\newblock \emph{arXiv preprint arXiv:2107.03374}, 2021.

\bibitem[Dabre et~al.(2021)Dabre, Imankulova, Kaneko, and Chakrabarty]{dabre2021simultaneousmultipivotneuralmachine}
R.~Dabre, A.~Imankulova, M.~Kaneko, and A.~Chakrabarty.
\newblock Simultaneous multi-pivot neural machine translation, 2021.
\newblock URL \url{https://arxiv.org/abs/2104.07410}.

\bibitem[DeepSeek-AI(2025)]{deepseekai2025deepseekr1incentivizingreasoningcapability}
DeepSeek-AI.
\newblock Deepseek-r1: Incentivizing reasoning capability in llms via reinforcement learning, 2025.
\newblock URL \url{https://arxiv.org/abs/2501.12948}.

\bibitem[Feyisetan et~al.(2022)Feyisetan, Ghanavati, Thaine, Habernal, and Mireshghallah]{privatenlp-2022-privacy-nlp}
O.~Feyisetan, S.~Ghanavati, P.~Thaine, I.~Habernal, and F.~Mireshghallah, editors.
\newblock \emph{Proceedings of the Fourth Workshop on Privacy in Natural Language Processing}, Seattle, United States, July 2022. Association for Computational Linguistics.
\newblock URL \url{https://aclanthology.org/2022.privatenlp-1.0}.

\bibitem[Haddow et~al.(2022)Haddow, Bawden, Miceli~Barone, Helcl, and Birch]{haddow-etal-2022-survey-mt}
B.~Haddow, R.~Bawden, A.~V. Miceli~Barone, J.~Helcl, and A.~Birch.
\newblock Survey of low-resource machine translation.
\newblock \emph{Computational Linguistics}, 48\penalty0 (3):\penalty0 673--732, Sept. 2022.
\newblock \doi{10.1162/coli_a_00446}.
\newblock URL \url{https://aclanthology.org/2022.cl-3.6}.

\bibitem[Hadi et~al.(2023)Hadi, Al-Tashi, Qureshi, Shah, Muneer, Irfan, Zafar, Shaikh, Akhtar, Al-Garadi, et~al.]{hadi2023large}
M.~U. Hadi, Q.~Al-Tashi, R.~Qureshi, A.~Shah, A.~Muneer, M.~Irfan, A.~Zafar, M.~B. Shaikh, N.~Akhtar, M.~A. Al-Garadi, et~al.
\newblock Large language models: A comprehensive survey of its applications, challenges, limitations, and future prospects.
\newblock 2023.

\bibitem[He et~al.(2020)He, Wang, Xiong, and Liu]{he-etal-2020-box}
J.~He, T.~Wang, D.~Xiong, and Q.~Liu.
\newblock The box is in the pen: Evaluating commonsense reasoning in neural machine translation.
\newblock In \emph{Findings of the Association for Computational Linguistics: EMNLP 2020}, pages 3662--3672, Online, Nov. 2020. Association for Computational Linguistics.
\newblock \doi{10.18653/v1/2020.findings-emnlp.327}.
\newblock URL \url{https://aclanthology.org/2020.findings-emnlp.327}.

\bibitem[He et~al.(2023)He, Liang, Jiao, Zhang, Yang, Wang, Tu, Shi, and Wang]{he2023exploring_maps_mt}
Z.~He, T.~Liang, W.~Jiao, Z.~Zhang, Y.~Yang, R.~Wang, Z.~Tu, S.~Shi, and X.~Wang.
\newblock Exploring human-like translation strategy with large language models.
\newblock \emph{arXiv preprint arXiv:2305.04118}, 2023.

\bibitem[He et~al.(2024)He, Wang, Jiao, Zhang, Wang, Shi, and Tu]{he2024improving_mt}
Z.~He, X.~Wang, W.~Jiao, Z.~Zhang, R.~Wang, S.~Shi, and Z.~Tu.
\newblock Improving machine translation with human feedback: An exploration of quality estimation as a reward model.
\newblock \emph{arXiv preprint arXiv:2401.12873}, 2024.

\bibitem[Hurst et~al.(2024)Hurst, Lerer, Goucher, Perelman, Ramesh, Clark, Ostrow, Welihinda, Hayes, Radford, et~al.]{hurst2024gpt4o}
A.~Hurst, A.~Lerer, A.~P. Goucher, A.~Perelman, A.~Ramesh, A.~Clark, A.~Ostrow, A.~Welihinda, A.~Hayes, A.~Radford, et~al.
\newblock Gpt-4o system card.
\newblock \emph{arXiv preprint arXiv:2410.21276}, 2024.

\bibitem[Jaech et~al.(2024)Jaech, Kalai, Lerer, Richardson, El-Kishky, Low, Helyar, Madry, Beutel, Carney, et~al.]{jaech2024openai_o1}
A.~Jaech, A.~Kalai, A.~Lerer, A.~Richardson, A.~El-Kishky, A.~Low, A.~Helyar, A.~Madry, A.~Beutel, A.~Carney, et~al.
\newblock Openai o1 system card.
\newblock \emph{arXiv preprint arXiv:2412.16720}, 2024.

\bibitem[Jiao et~al.(2023)Jiao, Wang, Huang, Wang, and Tu]{jiao2023chatgpt}
W.~Jiao, W.~Wang, J.-t. Huang, X.~Wang, and Z.~Tu.
\newblock Is chatgpt a good translator? a preliminary study.
\newblock \emph{arXiv preprint arXiv:2301.08745}, 2023.

\bibitem[Kim et~al.(2019)Kim, Petrov, Petrushkov, Khadivi, and Ney]{kim-etal-2019-pivot}
Y.~Kim, P.~Petrov, P.~Petrushkov, S.~Khadivi, and H.~Ney.
\newblock Pivot-based transfer learning for neural machine translation between non-{E}nglish languages.
\newblock In \emph{Proceedings of the 2019 Conference on Empirical Methods in Natural Language Processing and the 9th International Joint Conference on Natural Language Processing (EMNLP-IJCNLP)}, pages 866--876, Hong Kong, China, Nov. 2019. Association for Computational Linguistics.
\newblock \doi{10.18653/v1/D19-1080}.
\newblock URL \url{https://aclanthology.org/D19-1080}.

\bibitem[Klymenko et~al.(2022)Klymenko, Meisenbacher, and Matthes]{klymenko-etal-2022-differential}
O.~Klymenko, S.~Meisenbacher, and F.~Matthes.
\newblock Differential privacy in natural language processing the story so far.
\newblock In \emph{Proceedings of the Fourth Workshop on Privacy in Natural Language Processing}, pages 1--11, Seattle, United States, July 2022. Association for Computational Linguistics.
\newblock \doi{10.18653/v1/2022.privatenlp-1.1}.
\newblock URL \url{https://aclanthology.org/2022.privatenlp-1.1}.

\bibitem[Kocmi et~al.(2022)Kocmi, Bawden, Bojar, Dvorkovich, Federmann, Fishel, Gowda, Graham, Grundkiewicz, Haddow, Knowles, Koehn, Monz, Morishita, Nagata, Nakazawa, Nov{\'a}k, Popel, and Popovi{\'c}]{kocmi-etal-2022-findings}
T.~Kocmi, R.~Bawden, O.~Bojar, A.~Dvorkovich, C.~Federmann, M.~Fishel, T.~Gowda, Y.~Graham, R.~Grundkiewicz, B.~Haddow, R.~Knowles, P.~Koehn, C.~Monz, M.~Morishita, M.~Nagata, T.~Nakazawa, M.~Nov{\'a}k, M.~Popel, and M.~Popovi{\'c}.
\newblock Findings of the 2022 conference on machine translation ({WMT}22).
\newblock In \emph{Proceedings of the Seventh Conference on Machine Translation (WMT)}, pages 1--45, Abu Dhabi, United Arab Emirates (Hybrid), Dec. 2022. Association for Computational Linguistics.
\newblock URL \url{https://aclanthology.org/2022.wmt-1.1}.

\bibitem[Koehn and Knowles(2017)]{koehn2017six}
P.~Koehn and R.~Knowles.
\newblock Six challenges for neural machine translation.
\newblock \emph{arXiv preprint arXiv:1706.03872}, 2017.

\bibitem[Laskar et~al.(2023)Laskar, Bari, Rahman, Bhuiyan, Joty, and Huang]{laskar-etal-2023-systematic_eval_chatgpt}
M.~T.~R. Laskar, M.~S. Bari, M.~Rahman, M.~A.~H. Bhuiyan, S.~Joty, and J.~Huang.
\newblock A systematic study and comprehensive evaluation of {C}hat{GPT} on benchmark datasets.
\newblock In \emph{Findings of the Association for Computational Linguistics: ACL 2023}, pages 431--469, Toronto, Canada, July 2023. Association for Computational Linguistics.
\newblock \doi{10.18653/v1/2023.findings-acl.29}.
\newblock URL \url{https://aclanthology.org/2023.findings-acl.29}.

\bibitem[Li et~al.(2023{\natexlab{a}})Li, Koto, Wu, Aji, and Baldwin]{li2023bactrian}
H.~Li, F.~Koto, M.~Wu, A.~F. Aji, and T.~Baldwin.
\newblock Bactrian-x: A multilingual replicable instruction-following model with low-rank adaptation.
\newblock \emph{arXiv preprint arXiv:2305.15011}, 2023{\natexlab{a}}.

\bibitem[Li et~al.(2020)Li, Xu, Cui, Huang, Wei, Li, and Zhou]{li2020docbank}
M.~Li, Y.~Xu, L.~Cui, S.~Huang, F.~Wei, Z.~Li, and M.~Zhou.
\newblock Docbank: A benchmark dataset for document layout analysis, 2020.

\bibitem[Li et~al.(2023{\natexlab{b}})Li, Tan, and Liu]{li2023privacypreserving}
Y.~Li, Z.~Tan, and Y.~Liu.
\newblock Privacy-preserving prompt tuning for large language model services, 2023{\natexlab{b}}.

\bibitem[Lyu et~al.(2024)Lyu, Du, Xu, Duan, Wu, Lynn, Aji, Wong, and Wang]{lyu-etal-2024-paradigm-mt-llms}
C.~Lyu, Z.~Du, J.~Xu, Y.~Duan, M.~Wu, T.~Lynn, A.~F. Aji, D.~F. Wong, and L.~Wang.
\newblock A paradigm shift: The future of machine translation lies with large language models.
\newblock In N.~Calzolari, M.-Y. Kan, V.~Hoste, A.~Lenci, S.~Sakti, and N.~Xue, editors, \emph{Proceedings of the 2024 Joint International Conference on Computational Linguistics, Language Resources and Evaluation (LREC-COLING 2024)}, pages 1339--1352, Torino, Italia, May 2024. ELRA and ICCL.
\newblock URL \url{https://aclanthology.org/2024.lrec-main.120/}.

\bibitem[Moslem et~al.(2023)Moslem, Haque, and Way]{moslem2023adaptive}
Y.~Moslem, R.~Haque, and A.~Way.
\newblock Adaptive machine translation with large language models.
\newblock \emph{arXiv preprint arXiv:2301.13294}, 2023.

\bibitem[Niu and Carpuat(2020)]{Niu20controlling}
X.~Niu and M.~Carpuat.
\newblock Controlling neural machine translation formality with synthetic supervision.
\newblock \emph{Proceedings of the AAAI Conference on Artificial Intelligence}, 34\penalty0 (05):\penalty0 8568--8575, Apr. 2020.
\newblock \doi{10.1609/aaai.v34i05.6379}.
\newblock URL \url{https://ojs.aaai.org/index.php/AAAI/article/view/6379}.

\bibitem[Ouyang et~al.(2022)Ouyang, Wu, Jiang, Almeida, Wainwright, Mishkin, Zhang, Agarwal, Slama, Gray, et~al.]{ouyangtraining-instructGPT}
L.~Ouyang, J.~Wu, X.~Jiang, D.~Almeida, C.~Wainwright, P.~Mishkin, C.~Zhang, S.~Agarwal, K.~Slama, A.~Gray, et~al.
\newblock Training language models to follow instructions with human feedback.
\newblock In \emph{Advances in Neural Information Processing Systems}, 2022.

\bibitem[Pang et~al.(2024)Pang, Ye, Wang, Yu, Wong, Shi, and Tu]{pang2024salute}
J.~Pang, F.~Ye, L.~Wang, D.~Yu, D.~F. Wong, S.~Shi, and Z.~Tu.
\newblock Salute the classic: Revisiting challenges of machine translation in the age of large language models.
\newblock \emph{arXiv preprint arXiv:2401.08350}, 2024.

\bibitem[Qwen et~al.(2025)Qwen, :, Yang, Yang, Zhang, Hui, Zheng, Yu, Li, Liu, Huang, Wei, Lin, Yang, Tu, Zhang, Yang, Yang, Zhou, Lin, Dang, Lu, Bao, Yang, Yu, Li, Xue, Zhang, Zhu, Men, Lin, Li, Tang, Xia, Ren, Ren, Fan, Su, Zhang, Wan, Liu, Cui, Zhang, and Qiu]{qwen2025qwen25technicalreport}
Qwen, :, A.~Yang, B.~Yang, B.~Zhang, B.~Hui, B.~Zheng, B.~Yu, C.~Li, D.~Liu, F.~Huang, H.~Wei, H.~Lin, J.~Yang, J.~Tu, J.~Zhang, J.~Yang, J.~Yang, J.~Zhou, J.~Lin, K.~Dang, K.~Lu, K.~Bao, K.~Yang, L.~Yu, M.~Li, M.~Xue, P.~Zhang, Q.~Zhu, R.~Men, R.~Lin, T.~Li, T.~Tang, T.~Xia, X.~Ren, X.~Ren, Y.~Fan, Y.~Su, Y.~Zhang, Y.~Wan, Y.~Liu, Z.~Cui, Z.~Zhang, and Z.~Qiu.
\newblock Qwen2.5 technical report, 2025.
\newblock URL \url{https://arxiv.org/abs/2412.15115}.

\bibitem[Robinson et~al.(2023)Robinson, Ogayo, Mortensen, and Neubig]{robinson2023chatgpt_mt}
N.~R. Robinson, P.~Ogayo, D.~R. Mortensen, and G.~Neubig.
\newblock Chatgpt mt: Competitive for high-(but not low-) resource languages.
\newblock \emph{arXiv preprint arXiv:2309.07423}, 2023.

\bibitem[Sato and Nagao(1990)]{DBLP:conf/coling/SatoN90_MT}
S.~Sato and M.~Nagao.
\newblock Toward memory-based translation.
\newblock In \emph{13th International Conference on Computational Linguistics, {COLING} 1990, University of Helsinki, Finland, August 20-25, 1990}, pages 247--252, 1990.
\newblock URL \url{https://aclanthology.org/C90-3044/}.

\bibitem[Sennrich et~al.(2016)Sennrich, Haddow, and Birch]{Sennrich16controlling}
R.~Sennrich, B.~Haddow, and A.~Birch.
\newblock Controlling politeness in neural machine translation via side constraints.
\newblock In \emph{Proceedings of the 2016 Conference of the North {A}merican Chapter of the Association for Computational Linguistics: Human Language Technologies}, pages 35--40, San Diego, California, June 2016. Association for Computational Linguistics.
\newblock \doi{10.18653/v1/N16-1005}.
\newblock URL \url{https://www.aclweb.org/anthology/N16-1005}.

\bibitem[Stahlberg(2020)]{stahlberg2020neural}
F.~Stahlberg.
\newblock Neural machine translation: A review.
\newblock \emph{Journal of Artificial Intelligence Research}, 69:\penalty0 343--418, 2020.

\bibitem[Sulubacak et~al.(2020)Sulubacak, Caglayan, Gr{\"o}nroos, Rouhe, Elliott, Specia, and Tiedemann]{sulubacak2020multimodal}
U.~Sulubacak, O.~Caglayan, S.-A. Gr{\"o}nroos, A.~Rouhe, D.~Elliott, L.~Specia, and J.~Tiedemann.
\newblock Multimodal machine translation through visuals and speech.
\newblock \emph{Machine Translation}, 34:\penalty0 97--147, 2020.

\bibitem[Team(2024{\natexlab{a}})]{qvq-72b-preview}
Q.~Team.
\newblock Qvq: To see the world with wisdom, December 2024{\natexlab{a}}.
\newblock URL \url{https://qwenlm.github.io/blog/qvq-72b-preview/}.

\bibitem[Team(2024{\natexlab{b}})]{qwq-32b-preview}
Q.~Team.
\newblock Qwq: Reflect deeply on the boundaries of the unknown, November 2024{\natexlab{b}}.
\newblock URL \url{https://qwenlm.github.io/blog/qwq-32b-preview/}.

\bibitem[Touvron et~al.(2023)Touvron, Lavril, Izacard, Martinet, Lachaux, Lacroix, Rozière, Goyal, Hambro, Azhar, Rodriguez, Joulin, Grave, and Lample]{touvron2023llama1}
H.~Touvron, T.~Lavril, G.~Izacard, X.~Martinet, M.-A. Lachaux, T.~Lacroix, B.~Rozière, N.~Goyal, E.~Hambro, F.~Azhar, A.~Rodriguez, A.~Joulin, E.~Grave, and G.~Lample.
\newblock Llama: Open and efficient foundation language models, 2023.

\bibitem[Tsujii(1986)]{tsujii1986future_MT}
J.~Tsujii.
\newblock Future directions of machine translation.
\newblock In \emph{Coling 1986 Volume 1: The 11th International Conference on Computational Linguistics}, 1986.

\bibitem[Vaswani et~al.(2017)Vaswani, Shazeer, Parmar, Uszkoreit, Jones, Gomez, Kaiser, and Polosukhin]{Vaswani17attention}
A.~Vaswani, N.~Shazeer, N.~Parmar, J.~Uszkoreit, L.~Jones, A.~N. Gomez, L.~u. Kaiser, and I.~Polosukhin.
\newblock Attention is all you need.
\newblock In I.~Guyon, U.~V. Luxburg, S.~Bengio, H.~Wallach, R.~Fergus, S.~Vishwanathan, and R.~Garnett, editors, \emph{Advances in Neural Information Processing Systems 30}, pages 5998--6008. Curran Associates, Inc., 2017.
\newblock URL \url{http://papers.nips.cc/paper/7181-attention-is-all-you-need.pdf}.

\bibitem[Wang et~al.(2024{\natexlab{a}})Wang, Meng, Liang, and Zhou]{wang2024drt_o1}
J.~Wang, F.~Meng, Y.~Liang, and J.~Zhou.
\newblock Drt-o1: Optimized deep reasoning translation via long chain-of-thought.
\newblock \emph{arXiv preprint arXiv:2412.17498}, 2024{\natexlab{a}}.

\bibitem[Wang(2019)]{wang2019discourse}
L.~Wang.
\newblock \emph{Discourse-aware neural machine translation}.
\newblock PhD thesis, Dublin City University. School of Computing, 2019.

\bibitem[Wang et~al.(2023)Wang, Lyu, Ji, Zhang, Yu, Shi, and Tu]{wang2023document}
L.~Wang, C.~Lyu, T.~Ji, Z.~Zhang, D.~Yu, S.~Shi, and Z.~Tu.
\newblock Document-level machine translation with large language models.
\newblock \emph{arXiv preprint arXiv:2304.02210}, 2023.

\bibitem[Wang et~al.(2024{\natexlab{b}})Wang, Liu, Lyu, Jiao, Wang, Xu, Tu, Gu, Chen, Wu, et~al.]{wang2024findings_wmt_24}
L.~Wang, S.~Liu, C.~Lyu, W.~Jiao, X.~Wang, J.~Xu, Z.~Tu, Y.~Gu, W.~Chen, M.~Wu, et~al.
\newblock Findings of the wmt 2024 shared task on discourse-level literary translation.
\newblock \emph{arXiv preprint arXiv:2412.11732}, 2024{\natexlab{b}}.

\bibitem[Wang et~al.(2022)Wang, Sun, Cheng, Zheng, and Wang]{wang2022controlling}
Y.~Wang, Z.~Sun, S.~Cheng, W.~Zheng, and M.~Wang.
\newblock Controlling styles in neural machine translation with activation prompt.
\newblock \emph{arXiv preprint arXiv:2212.08909}, 2022.

\bibitem[Wei et~al.(2022{\natexlab{a}})Wei, Bosma, Zhao, Guu, Yu, Lester, Du, Dai, and Le]{weifinetuned}
J.~Wei, M.~Bosma, V.~Zhao, K.~Guu, A.~W. Yu, B.~Lester, N.~Du, A.~M. Dai, and Q.~V. Le.
\newblock Finetuned language models are zero-shot learners.
\newblock In \emph{International Conference on Learning Representations}, 2022{\natexlab{a}}.

\bibitem[Wei et~al.(2022{\natexlab{b}})Wei, Wang, Schuurmans, Bosma, Xia, Chi, Le, Zhou, et~al.]{wei2022chain_cot}
J.~Wei, X.~Wang, D.~Schuurmans, M.~Bosma, F.~Xia, E.~Chi, Q.~V. Le, D.~Zhou, et~al.
\newblock Chain-of-thought prompting elicits reasoning in large language models.
\newblock \emph{Advances in neural information processing systems}, 35:\penalty0 24824--24837, 2022{\natexlab{b}}.

\bibitem[Wu and Wang(2007)]{wu2007pivot}
H.~Wu and H.~Wang.
\newblock Pivot language approach for phrase-based statistical machine translation.
\newblock \emph{Machine Translation}, 21:\penalty0 165--181, 2007.

\bibitem[Xia et~al.(2025)Xia, Li, Leong, Wang, and Li]{xia2025tokenskipcontrollablechainofthoughtcompression}
H.~Xia, Y.~Li, C.~T. Leong, W.~Wang, and W.~Li.
\newblock Tokenskip: Controllable chain-of-thought compression in llms, 2025.
\newblock URL \url{https://arxiv.org/abs/2502.12067}.

\bibitem[Yang et~al.(2020)Yang, Wang, and Chu]{yang2020survey-mt}
S.~Yang, Y.~Wang, and X.~Chu.
\newblock A survey of deep learning techniques for neural machine translation, 2020.

\bibitem[Zhang et~al.(2025)Zhang, Chen, Bai, Li, Xiang, and Zhang]{zhang2025exploringtranslationmechanismlarge}
H.~Zhang, K.~Chen, X.~Bai, X.~Li, Y.~Xiang, and M.~Zhang.
\newblock Exploring translation mechanism of large language models, 2025.
\newblock URL \url{https://arxiv.org/abs/2502.11806}.

\bibitem[Zhao et~al.(2024)Zhao, Yin, Zeng, Wang, Shi, Lyu, Wang, Luo, and Zhang]{zhao2024marco_o1}
Y.~Zhao, H.~Yin, B.~Zeng, H.~Wang, T.~Shi, C.~Lyu, L.~Wang, W.~Luo, and K.~Zhang.
\newblock Marco-o1: Towards open reasoning models for open-ended solutions.
\newblock \emph{arXiv preprint arXiv:2411.14405}, 2024.

\end{thebibliography}


\end{CJK}
\end{document}